\documentclass{article}
\usepackage{natbib}%
\usepackage[figuresright]{rotating}

\usepackage{url}
\usepackage{algorithmic}
\usepackage{algorithm}

\usepackage{graphicx,subfigure}
\usepackage{multicol}
\usepackage{multirow}
\usepackage{amsmath,amsthm}

\usepackage{hyperref} 
\theoremstyle{definition}

\usepackage[table]{xcolor}
\definecolor{red}{RGB}{255, 0, 0}
\theoremstyle{remark}

\begin{document}
	
	\title{Improving a State-of-the-Art Heuristic for the Minimum Latency Problem with Data Mining}

        \author {
Ítalo Santana\textsuperscript{\rm 1},
Alexandre Plastino\textsuperscript{\rm 1},
and Isabel Rosseti\textsuperscript{\rm 1},\\\\
\textsuperscript{\rm 1}Instituto de Computa\c{c}\~{a}o, Universidade Federal Fluminense,\\Av. Gal. Milton Tavares de Souza, s/n, Niter\'{o}i, 24210-346, Brazil\\\\	
 	isantana@id.uff.br; plastino@ic.uff.br; rosseti@ic.uff.br
}

\date{}
	
	\maketitle

	
	
	
	
	
	\begin{abstract}
		Recently, hybrid metaheuristics have become a trend in operations research. A successful example combines the Greedy Randomized Adaptive Search Procedures (GRASP) and data mining techniques, where frequent patterns found in high-quality solutions can lead to an efficient exploration of the search space, along with a significant reduction of computational time. In this paper, a GRASP-based state-of-the-art heuristic for the Minimum Latency Problem is improved by means of data mining techniques. Computational experiments showed that the hybrid heuristic with data mining was able to match or improve the solution quality for a large number of instances, together with a substantial reduction of running time. Besides, 32 new best known solutions are introduced to the literature. To support our results, statistical significance tests, analyses over the impact of mined patterns, comparisons based on running time as stopping criterion, and time-to-target plots are provided.
	\end{abstract}
	
	
	
	\section{Introduction}

	Metaheuristics are widely used to solve combinatorial optimization problems, providing near-optimal solutions in practical running time. Proposed in~\cite{Feo1995}, Greedy Randomized Adaptive Search Procedures (GRASP) is a well-known metaheuristic successfully applied to several combinatorial optimization problems. Briefly, GRASP is a simple iterative process composed of a constructive phase and a local search phase. Its mechanism works, at each iteration, constructing a feasible solution by a semi-greedy procedure to, then, be improved by exploring its neighbor search space.
	
	Combining components from different metaheuristics, even from distinct aspects and paradigms, has become a trend in combinatorial optimization, hence, named as hybrid metaheuristics \citep{Talbi2002}. To extend their efficiency and flexibility, hybrid metaheuristics are also able to incorporate concepts from other fields of study, such as Data Mining (DM). DM refers to the automatic extraction of knowledge from datasets, which can be represented by means of rules or patterns \citep{Han2011}.
	
	In heuristics, the underlying idea of incorporating DM techniques consists in extracting patterns from good solutions to guide the search procedure. For instance, a hybrid approach based on GRASP with DM techniques, named Data Mining GRASP (DM-GRASP), was presented in \cite{Ribeiro2004, Ribeiro2006} for the Set Packing Problem. The basic concept of DM-GRASP consists in extracting patterns from high-quality solutions to, in a subsequent phase, use mined patterns to construct initial solutions. By their promising results, this approach was also applied to several problems, such as the Maximum Diversity Problem \citep{Santos2005}, the Server Replication for Reliable Multicast Problem \citep{Santos2006}, the p-Median Problem \citep{Plastino2011}, the 2-Path Network Design Problem \citep{Barbalho2013} and the One-Commodity Pickup-and-Delivery Traveling Salesman Problem \citep{Guerine2016}.
	
	Afterward, an adaptive version of DM-GRASP, named Multi DM-GRASP (MDM-GRASP), was proposed in \cite{Plastino2011} for the p-Median Problem. Different from the DM-GRASP strategy, where the data mining process is performed once, the approach presented in MDM-GRASP performs the data mining process whenever the elite set becomes stable. This stability refers to a given number of iterations without any changes in the elite set. In practical terms, the general quality of the elite set is progressively enhanced throughout the execution, yielding, at each time, more refined patterns. Indeed, the results from the MDM version were superior when compared to those from the original heuristics and reasonably better than those from the DM version \citep{Plastino2011, Barbalho2013, Plastino2014, Guerine2016}, both in terms of solution quality and computational time.
	
	In this paper, we propose a hybrid data mining approach inspired by MDM-GRASP for solving the Minimum Latency Problem (MLP). The objective of MLP, a Traveling Salesman Problem (TSP) variant, consists in finding a Hamiltonian Circuit that minimizes the arrival times in each vertex, which is seen as a customer-centric routing problem. Real-life examples of the MLP are easily found in logistic planning \citep{Fischetti1993,Blum1994,Campbell2008}.
	
	In our proposed hybrid DM strategy, we hybridized a state-of-the-art heuristic called GILS-RVND, which was presented in \cite{Silva2012} for the MLP. Computational results obtained in our experiments showed that our hybrid data mining approach, called MDM-GILS-RVND, can solve more efficiently virtually all 173 instances originally reported in~\cite{Silva2012}, not only in solution quality but also in computational time. Beyond that, in order to extend our experiments, a 56-instances set is introduced to this paper, composed of all instances from 120 to 1379 customers found in TSPLib\footnote{Instances collected from http://elib.zib.de/pub/mp-testdata/tsp/tsplib/tsp/index.html}. This new set aims at better evaluating MDM-GILS-RVND and GILS-RVND. In total, 229 instances were tested, and their results are reported in this paper.
	
	It is noteworthy that the original source code of GILS-RVND as well as all seeds of pseudo-random numbers used in the experiments reported in \cite{Silva2012} were provided by the authors, which allowed us to entirely re-execute the original experiments of this heuristic, providing, in this way, a fair comparison between the heuristics evaluated in this paper. Furthermore, to support all computational experiments, statistical significance tests, analyses over the impact of mined patterns, experiments using computational time as the stopping criterion, and time-to-target plots are presented. Finally, we also made available all log files of all reported experiments on an online repository \citep[see][]{Santana2018dataset}.
	
	From these results and analyses, MDM-GILS-RVND demonstrated to be more efficient than GILS-RVND in terms of solution quality and computational time simultaneously, mainly in challenging instances, which usually start from 500 customers in this work. Besides, the proposed method was able to find better solutions for 32 instances, adding 32 best known solutions to the literature.
	
	The remaining of this paper is organized as follows. In Section~\ref{sec:mlp}, the Minimum Latency Problem is introduced along with its related literature. Section~\ref{sec:gils-rvnd} presents the GILS-RVND heuristic. In Section~\ref{sec:mdm-gils-rvnd}, the hybrid heuristic with data mining is presented. Computational evaluation comparing both heuristics is displayed in Section~\ref{sec:computational-experiments}. Concluding remarks and future works are pointed out in Section~\ref{sec:conclusion}.

	\section{The Minimum Latency Problem} \label{sec:mlp}
	
	In this section, the Minimum Latency Problem (MLP) is defined and relevant methods to solve this problem are reviewed, respectively, in Subsections 2.1 and 2.2.
	
	\subsection{Problem Definition} \label{subsec:problem-description}
	
	Let $G = (V,A)$ be a complete directed graph, where $V = \{0, \dots, n\}$ stands for the set of vertices, being~$0$ the depot and the remainder ones representing the customers, and $A = \{(i,j):i,j \in V, i \neq j\}$ is the set of arcs, where each one is associated with a travel time between $i$~and~$j$. The MLP aims at finding a Hamiltonian circuit ($i_0, i_1,\ldots,i_n, i_{n+1}$) in $G$, which starts and ends at the depot ($i_0=0$ and $i_{n+1}=0$), that minimizes $\sum_{k=1}^{n+1}l(i_k)$, where $l(i_k)$ is the accumulated travel time from the depot to $i_k$.
	
\begin{figure}[htbp]
		\centering
		\subfigure[Solution~$s$]{
			\label{fig:solucao_viavel}
			\includegraphics[scale=0.45]{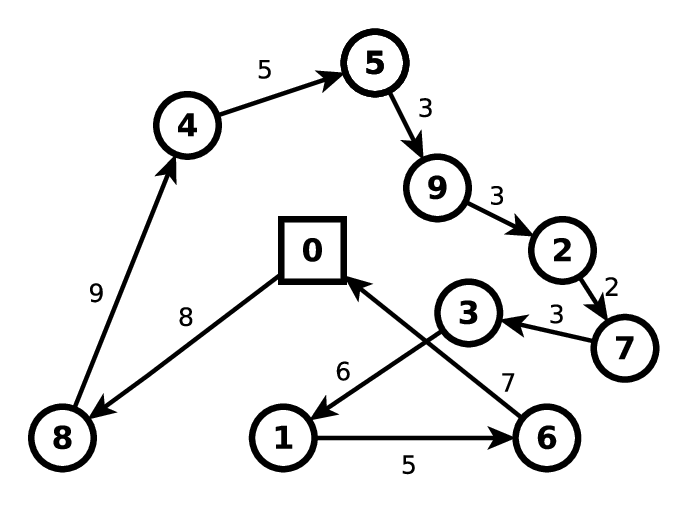}
		}
		\subfigure[Hamiltonian circuit and Hamiltonian path variants have $f(s)$ equal to, respectively, 297 and 246]{
			\label{fig:custos_acumulados}
			\includegraphics[scale=0.55]{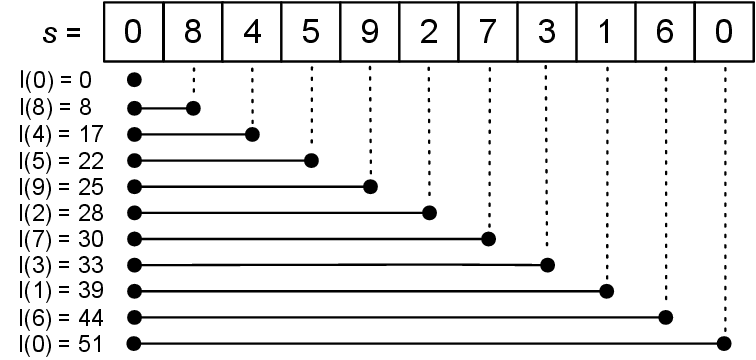}
		}
		\caption{An MLP feasible solution}
		\label{fig:feasible-solution-example}
	\end{figure}

	As an example, consider $s$ as an MLP feasible solution presented in Fig.~\ref{fig:solucao_viavel} and its sequence of visited customers in Fig.~\ref{fig:custos_acumulados}. Thus, the total latency (cost value) of $s$, or $f(s)$, is 297.
	
	Besides the Hamiltonian circuit definition, \cite{Salehipour2011} considered an MLP variant whose objective is to find a Hamiltonian path ($i_0, i_1,\ldots,i_n$) in $G$, which starts at the depot ($i_0=0$), that minimizes $\sum_{k=1}^{n}l(i_k)$. In Fig.~\ref{fig:custos_acumulados}, the $f(s)$ for this variant corresponds to 246.
	
	Despite its simple definition, the MLP was proven $\mathcal{NP}$-hard for general metric spaces in~\cite{Sahni1976}. The MLP is a Traveling Salesman Problem (TSP) variant, and it is known under different names: Traveling Repairman Problem \citep{Afrati1986}, Delivery Man Problem \citep{Fischetti1993}, Cumulative Traveling Salesman Problem \citep{Bianco1993} and School Bus Driver Problem \citep{Chaudhuri2003}. In a general context, MLP is recognized being much harder than TSP, since minor changes in the sequence of visited customers in an MLP solution can propagate major non-local changes in its general structure~\citep{Blum1994}.
	
	Many real-life applications are associated with MLP due to its customer viewpoint aspect, different from the server viewpoint aspect present in TSP. Several examples of MLP instances can be found in daily life, such as delivery of goods \citep{Fischetti1993}, disk-head scheduling \citep{Blum1994}, disaster situations \citep{Campbell2008}, and data retrieval in networks \citep{Ezzine2010}.
	
	\subsection{Literature Review} \label{subsec:literature-review}
	
	In this subsection, we review relevant approaches for the MLP, which are divided into exact algorithms, approximation algorithms, and heuristics.
	
	Many exact algorithms have been proposed in the literature for the MLP. In \cite{Lucena1990}, a branch-and-bound algorithm based on a lower bound scheme was developed and tested on instances containing between 15 and 30 nodes. A matroidal structure to obtain lower bounds for an enumerative algorithm was used in~\cite{Fischetti1993}, involving instances including up to 60 vertices. Two exact algorithms that incorporate lower bounds provided by Lagrangian relaxations were reported in \cite{Bianco1993}. In \cite{Eijl1995} and also in \cite{Mendez-Diaz2008}, new mixed integer programming formulations were presented, where lower bounds were obtained by linear programming relaxations. Later on, two new linear integer formulations, based on previous formulations found in the literature, were proposed in~\cite{Ezzine2010}, which could deal with instances with no more than 29 vertices. A branch-and-cut-and-price algorithm, introduced in \cite{Abeledo2010a,Abeledo2010b}, succeed to handle instances with 107 vertices. Based on branch-and-bound, there are also two formulations for the MLP with time windows approach \citep{Heilporn2010}. One formulation is based on a classical arc flow model, and the other is presented as a sequential assignment model. Also, two new integer formulations, considering asymmetrical instances, were developed in \cite{Angel-bello2012}. \cite{Roberti2014} proposed a method based on column generation and dynamic programming, called \textit{dynamic ng-path relaxation}, which could solve instances including up to 150 vertices. \cite{Bulhoes2018} developed a branch-and-price algorithm combined with \textit{ng-path relaxation} capable of solving instances with 195 vertices to optimality.
	
	Approximation algorithms also appear in the literature for the MLP. These algorithms are guaranteed to find feasible solutions, in polynomial time, no worse than an approximation ratio in relation to the optimal solution. Reported in \cite{Blum1994}, the proposed approximation algorithm reached a ratio of 144. Later on, known for their relevancy, approximation algorithms with ratios of 3.59 and 3.03 were reported in \cite{Chaudhuri2003} and \cite{Archer2010}, respectively.
	
	Finally, several important heuristics have been proposed to solve the MLP. Apart from an enumerative algorithm, a heuristic approach was also designed in~\cite{Fischetti1993}. Heuristics for MLP and TSP were introduced in~\cite{Campbell2008}, where the reported experiments used different objective functions. In~\cite{Ngueveu2010}, a memetic algorithm to obtain upper bounds was developed. Besides two exact formulations for the MLP with time windows proposed in \cite{Heilporn2010}, two heuristics with and without a tabu list component were developed. Two GRASP-based heuristics with Variable Neighborhood Search (VNS) and another one with Variable Neighborhood Descent (VND) were elaborated and presented in~\cite{Salehipour2011}. Developed and published in~\cite{Silva2012}, the GILS-RVND is a simple and effective algorithm for the MLP based on GRASP, ILS (Iterated Local Search -- \citealt{Lourencco2003}), and RVND (Variable Neighborhood Descent with random neighborhood ordering -- \citealp{Mladenovic1997,Subramanian2010}). This heuristic holds nowadays the best results for a set of 173 instances, including instances with 1000 customers, being 23 instances tested in~\cite{Abeledo2010a,Abeledo2010b} and 150 instances from~\cite{Salehipour2011}. One year later, these instances were also adopted in~\cite{Mladenovic2013} to test two General Variable Neighborhood Search (GVNS) heuristics based on different strategies for the local search phase. Results reported in~\cite{Mladenovic2013} were only compared to results reported in works that introduced these instances \citep{Abeledo2010a, Abeledo2010b, Salehipour2011}, where some improvement on these instances was observed. Although the GVNS heuristics and GILS-RVND were not directly compared, analysing their reported results, GILS-RVND presented a general better performance in terms of solution quality. Thus, based on this evidence, GILS-RVND can be considered as a state-of-the-art heuristic for the MLP. Recently, parallel versions based on GILS-RVND were also proposed in~\cite{Rios2016}.
	
	\section{GILS-RVND: the State-of-the-Art Algorithm} \label{sec:gils-rvnd}
	
	The GILS-RVND is an approach developed by \cite{Silva2012} for solving two MLP variants: Hamiltonian circuit and Hamiltonian path. For tackling the latter variant, the latency of returning to the depot in a Hamiltonian circuit solution is ignored, making GILS-RVND able to solve these variants with a little adaptation in the algorithm.
	
	This heuristic was built on components of Greedy Randomized Adaptive Search Procedures (GRASP), Iterated Local Search (ILS), and Variable Neighborhood Descent with random neighborhood ordering (RVND). Beyond these components, GILS-RVND also implements a framework capable of evaluating moves of neighborhood structures in $O(1)$ amortized operations.
	
	Results reported in \cite{Silva2012} demonstrated that this heuristic found optimal solutions for instances including up to 50 customers in less than one second, and it was very effective in reaching high-quality solutions for the other instances whose sizes involve up to 1000 customers. Since GILS-RVND outperformed any existing heuristic at that time, its results opened new challenges for the MLP literature. Currently, this heuristic remains competitive when compared to recent approaches, which makes it be a state-of-the-art heuristic for the MLP.
	
	In Subsection~\ref{subsec:componentsGILS}, we detail the components in which GILS-RVND was built. Next, Subsection~\ref{subsec:pseudocodeGILS} shows the structure of GILS-RVND, where both constructive and local search procedures are described.
	
	\subsection{Components of GILS-RVND} \label{subsec:componentsGILS}
	
	In \cite{Feo1995}, GRASP is presented as a simple iterative process composed of two phases in each iteration, a constructive phase, and a local search phase. This iterative process ends when the stopping criterion is met, and then, the best solution found is returned. Responsible for generating feasible initial solutions, the constructive phase of GRASP uses a random component to control its greediness, allowing distinct starts for the local search phase at each iteration. In the constructive phase, a partial solution (initially empty) is built by inserting an element at a time until it becomes complete. The initial solution is probably not near-optimal, so, for its improvement, it is submitted to the second phase of GRASP, the local search phase. This phase explores the search space of the problem, in order to replace the current solution with a better solution considering the neighborhood. When neighbor solutions can not further improve the current solution, the local search phase terminates. GRASP terminates when its stopping criterion is reached, such as a predefined number of iterations.

	Iterated Local Search (ILS) is a simple and flexible metaheuristic for optimization problems \citep{Lourencco2003}. This metaheuristic builds its initial solution once, and submit it to an iterative process composed of a local search phase, a perturbation mechanism, and an acceptance criterion for solutions. Considering a basic ILS, the initial solution~$s$ can be entirely random or constructed by a semi-greedy procedure. Next, the loop of the ILS starts with the local search component working to improve $s$ by browsing in its neighborhood. After the local search, $s$  is submitted to a perturbation mechanism, allowing to avoid the drawback of being trapped into local optima that may happen to occur during the algorithm execution. This mechanism modifies the solution in such a way to escape from a local optimum, hence, leading the local search phase to explore another search space area. Finally, ILS uses an acceptance criterion to replace the best global solution by $s$, where $s$ is chosen not only by the usual cost value factor but also by factors involving random acceptance and diversification, as suggested in~\cite{Lourencco2003}. The iterative process of ILS terminates when the stopping criterion of the loop is reached, returning the best global solution found so far.
	
	VND with random neighborhood ordering (RVND) \citep{Subramanian2010}, which is a variant of the Variable Neighborhood Descent (VND) method \citep{Mladenovic1997}, performs the local search of GILS-RVND. VND method works with a list $N$ of all neighborhood structures sorted in a predefined order. Whenever a new better cost solution is found, this method restarts to the first neighborhood structure of $N$ at this new solution. Otherwise, the next neighborhood structure of $N$ is selected to continue the search. VND ends when all neighborhood structures fail to improve the same solution. In RVND, the VND is executed considering a randomized selection of neighborhood structures.
	
	Furthermore, a notable contribution included in GILS-RVND is related to a simple evaluation framework that requires $O(1)$ amortized operations to evaluate neighbor solutions. This approach was introduced in \cite{Kindervater1997} and extended in \cite{Vidal2015} for different Vehicle Routing Problem variants. This framework allows calculating cost values of neighbor solutions using preprocessed data structures, which turns to be faster than the traditional evaluation of considering all customers one by one. At first, a solution $s$ is decomposed into many subsequences, and each subsequence cost value is stored in specific data structures. Next, when the neighborhood of $s$ is explored for better solutions, each neighbor solution can be reached using these data structures, where a specific order of subsequences will generate a neighbor solution. Each specific order is associated with a neighborhood structure. Since five types of classic TSP neighborhood structures were used in \cite{Silva2012}, five different orders of subsequences were defined. Therefore, the use of preprocessed cost values of subsequences to calculate cost values of neighbor solutions made GILS-RVND a very efficient algorithm.
	
	\subsection{The Structure of GILS-RVND} \label{subsec:pseudocodeGILS}
	
	The pseudocode of GILS-RVND is detailed in Algorithm \ref{alg:gils-rvnd}. Firstly, the cost of the current best solution $s^*$ is initialized (line 1). In the outer loop, for each one of the $I_{Max}$ iterations, an initial solution $s$ is generated by a constructive procedure requiring an $\alpha \in R$ to control the greediness level of this procedure (line 3). Given as input data, the R set contains $I_{Max}$ random values from the interval $[0,1)$. Before starting the inner loop, $s$ is copied to $s'$ and $iterILS$ counter is set to zero (lines 4-5). From this point, $s'$ refers to the best solution of the local search phase (lines 6-14). The basic idea of this phase consists in exploring the neighborhood of $s$ using an RVND procedure (line 7) and a perturbation mechanism over $s'$ (line 12) until $I_{ILS}$ consecutive attempts of improvement without success in $s'$. This perturbation helps the local search from being trapped into local optima, using a move known as Double-bridge~\citep{Martin1991}, which was firstly proposed for the TSP. If the cost value of $s$, or $f(s)$, is lower than $f(s')$ (as MLP is a minimization problem), then, $s'$ is updated and $iterILS$ is reset. Next, $s'$ is perturbed and $iterILS$ is increased by one unit (lines 12-13). After the local search phase, $s^*$ may be updated (lines 15-17). Once all $I_{Max}$ iterations are performed, $s^*$ is returned (line 19).

	\begin{algorithm}
		\caption{GILS-RVND$(I_{Max}, I_{ILS}, R)$}\label{alg:gils-rvnd}
		\begin{algorithmic}[1]
			\STATE $f(s^*) \leftarrow \infty;$
			\FOR{$i = 1, \dots,  I_{Max}$ }
			\STATE $s \leftarrow ConstructiveProcedure(\alpha \in R);$
			\STATE $s' \leftarrow s;$
			\STATE $iterILS \leftarrow 0;$
			\WHILE{$iterILS < I_{ILS}$}
			\STATE $s \leftarrow RVND(s);$
			\IF{$f(s) < f(s')$}
			\STATE $s' \leftarrow s;$
			\STATE $iterILS \leftarrow 0;$
			\ENDIF
			\STATE $s \leftarrow Perturb(s');$
			\STATE $iterILS \leftarrow iterILS + 1;$
			\ENDWHILE
			\IF{$ f(s') < f(s^*) $}
			\STATE $s^* \leftarrow s';$
			\ENDIF
			\ENDFOR
			\STATE \textbf{return} $s^*$
		\end{algorithmic}
	\end{algorithm}
	
	In order to show how initial solutions are generated in GILS-RVND, its constructive procedure is explained in Algorithm \ref{alg:gils-rvnd-constructive}. At first, the depot (vertex 0) is straightaway inserted to the partial solution $s$ while the remaining vertices fill the Customer List $CL$ (lines 1-2). For simplification purposes, a variable $r$ is declared to always hold the last customer included in $s$ throughout this algorithm execution, where it receives the depot as initial value (line 3). Next, a systematic process to fill $s$ is executed in a loop (lines~4-11).  At each iteration, the customers of $CL$ are firstly sorted in ascending order by the travel time between each one and $r$ (line 5). After this sorting, the best ($\alpha$ x 100)\% customers from the $CL$ are copied to the Restricted Customer List $RCL$, where the less is $\alpha$, the shorter is $RCL$, and $\alpha = 0$ stands for the best customer of $RCL$ (line 6). Since $RCL$ gets filled, a random customer from this list, or $c$, is selected to, then, be inserted at the end of $s$ (lines 7-8). Next, the $r$ variable is updated and $c$ is removed from $CL$ (lines 9-10). Eventually, all elements from $CL$ are transferred to $s$, turning the partial solution into a feasible complete solution returned by the algorithm (line 12).
	
	\begin{algorithm}
		\caption{ConstructiveProcedure$(\alpha)$}\label{alg:gils-rvnd-constructive}
		\begin{algorithmic}[1]
			\STATE $ s \leftarrow \{0\};$
			\STATE $CL \leftarrow GenerateCL();$
			\STATE $r \leftarrow 0 ;$
			\WHILE{$CL \neq \emptyset $}
			\STATE $CL \leftarrow SortCL(CL,r);$
			\STATE $RCL \leftarrow FillRCL(CL,\alpha);$
			\STATE $c \leftarrow SelectRandomClient(RCL);$
			\STATE $s \leftarrow s \cup \{c\};$
			\STATE $r \leftarrow c;$
			\STATE $CL \leftarrow CL - \{c\}$
			\ENDWHILE
			\STATE \textbf{return} $s$
		\end{algorithmic}
	\end{algorithm}
	
	Next, the RVND procedure used in the local search of GILS-RVND is described. In a general view, RVND employs a set of neighborhood structures $NS$ as follows. This procedure randomly selects a neighborhood structure $nh \in NS$ to try the improvement of the current best solution~$s$. If $s$ is not improved by $nh$, then $nh$ is removed from $NS$. Otherwise, when $s$ is improved by $nh$, $s$ is updated and $NS$ is restored to its original state. The procedure goes on until all neighborhood structures fail to improve $s$, when $s$ is returned. GILS-RVND adopts five types of TSP traditional neighborhood structures in its local search phase:
	
	\begin{itemize}
		\item Swap: two customers are interchanged.
		\item 2-opt: two non-adjacent arcs are removed and two others are added, accepting only feasible solutions.
		\item Reinsertion: one customer is relocated to another position.
		\item Or-opt-2: one arc is relocated to another position.
		\item Or-opt-3: two adjacent arcs are relocated to another position.
	\end{itemize}
	
	Algorithm \ref{alg:gils-rvnd-rvnd} details the RVND procedure. At first, a neighborhood structures list $NS$ with all neighborhood structures and the data structures on subsequences of $s$ are initialized (lines 1-2). These data structures are components of the evaluation framework, which divides $s$ into many subsequences to evaluate its neighborhood in $O(1)$ amortized operations. In the inner loop (lines 3-13), a neighborhood structure $nh$ is randomly selected from $NS$ (line 4) to try the improvement of $s$ (line 5). In the neighborhood search of $nh$, the best improvement strategy is applied, which consists in moving $s$ to the best of its neighbors, whenever the latter improve the former. The resulting solution of this strategy is represented by $s'$. If $s$ is improved by $s'$, then $s$ is updated, $NS$ is reset, and the data structures on subsequences are updated using $s$ as reference (lines 6-9), otherwise, the current neighborhood structure is removed from $NS$ (lines 10-11). Once all neighborhood structures fail to improve $s$, this solution is returned (line~14).
	
	\begin{algorithm}
		\caption{RVND$(s)$}\label{alg:gils-rvnd-rvnd}
		\begin{algorithmic}[1]
			\STATE $ NS \leftarrow SetNeighborhoodList();$
			\STATE $ InitializeDataStructuresOnSubsequences(s); $
			\WHILE{$NS \neq \emptyset $}
			\STATE $nh \leftarrow SelectRandomNeighborhood(NS);$
			\STATE $s' \leftarrow FindBestNeighborSolution(nh,s);$
			\IF{$f(s') < f(s)$}
			\STATE $s \leftarrow s';$
			\STATE $NS \leftarrow SetNeighborhoodList();$
			\STATE $ UpdateDataStructuresOnSubsequences(s); $
			\ELSE
			\STATE $ NS \leftarrow NS - \{nh\}$;
			\ENDIF
			\ENDWHILE
			\STATE \textbf{return} $s$
		\end{algorithmic}
	\end{algorithm}
	
	\section{MDM-GILS-RVND: The Hybrid Heuristic with Data Mining} \label{sec:mdm-gils-rvnd}
	
	This section aims at presenting the process of incorporating data mining techniques into GILS-RVND. In Subsection~\ref{subsec:data-mining-concepts}, the Frequent Itemset Mining technique, which bases our proposed hybrid heuristic, is introduced. Next, Subsection~\ref{subsec:hybridization-heuristics-mdm} demonstrates how these concepts are incorporated into GRASP, presenting a general framework of hybridization seen so far in the literature. Finally, Subsection~\ref{subsec:hybridization-gils-rvnd-mdm} details the hybrid heuristic with data mining proposed in this work.
	
	\subsection{Frequent Itemset Mining} \label{subsec:data-mining-concepts}
	
	In a dataset composed of transactions, where each transaction corresponds to a set of elements from an application domain, relationships among data can be mined in terms of frequent itemsets (patterns). An itemset, which is a subset of items of the application domain, is mined if its support, a percentage indicator of its occurrence in the dataset, is greater or equal to a given minimum support. In this case, this subset of items is called a frequent itemset. Therefore, the mining of frequent itemsets consists in identifying all frequent itemsets in the dataset regarding a given minimum support. Besides that, maximal frequent itemsets can also be mined, where a maximal frequent itemset corresponds to a frequent itemset that is not a subset of any other frequent itemset.
	
	Having these concepts stated, let $E$$=$$\{e_1, e_2, e_3, \ldots , e_n\}$ be a set of the elements from the application domain. A transaction~$t$ is a subset of $E$ and a dataset~$D$ is a set of transactions. An itemset~$F \subseteq E$ with support $sup$, which is the percentage of transactions in $D$ where $F$ occurs, is said to be frequent when $sup$ is greater or equal than a given minimum support ($sup_{min}$). Thus, identifying all frequent itemsets in $D$ with $sup_{min}$ specified as a parameter is the Frequent Itemset Mining (FIM) problem, where several efficient algorithms have been proposed, such as Apriori \citep{Agrawal1994} and FP-Growth \citep{Han2000}. The algorithm FPMax* was proposed to mine efficiently maximal frequent itemsets \citep{Grahne2003}.
	
	For exemplification purposes, let $E$$=$$\{1, 2, 3, 4, 5\}$ be the set of all elements of the application domain and $sup_{min}$ equal to 80\%. Consider a dataset~$D$ composed of five transactions, e.g., $D$$=$$\{ \{1, 2, 3, 4, 5\}, \{1, 2, 3, 5\}, \{2, 3, 4, 5\}, \{1, 2, 5\}, \{2, 3, 4\}\}$.   By applying an FIM technique in $D$, the frequent itemsets extracted are $\{2\}$, $\{3\}$, $\{5\}$, $\{2,3\}$ and $\{2,5\}$, since they occur in, at least, 80\% of all transactions of~$D$. Note that, in case of mining maximal frequent itemsets, only $\{2,3\}$ and $\{2,5\}$ are extracted as these itemsets are not subsets of any frequent itemset.
	
	\subsection{DM-GRASP and MDM-GRASP} \label{subsec:hybridization-heuristics-mdm}
	
	The process of incorporating data mining techniques into heuristics comes from the idea of extracting patterns from near-optimal solutions, which can be used to lead the search for better solutions. The first approach of hybridizing these techniques into GRASP was proposed in~\cite{Ribeiro2004, Ribeiro2006}, being named as Data Mining GRASP (DM-GRASP). This approach consists in a two-phase GRASP separated by a data mining process. In the first phase, which lasts for a significant number of GRASP iterations, a set of different solutions is generated, storing the $d$ best solutions in the dataset $D$ (elite set), where each solution of $D$ represents a transaction $t$. Elite set, in this case, can be considered as a long term memory added to the algorithm.
	
	After the first phase, the data mining process, which is an application of an FIM technique, extracts a list of patterns~$P$ from $D$ using a predefined $sup_{min}$. In other words, a pattern $p \in P$ is a frequent itemset found in the solutions (transactions) of $D$ with support greater or equal than $sup_{min}$. Next, the second phase of DM-GRASP executes the remaining GRASP iterations, where, at each iteration, a solution based on a pattern $p \in P$ is constructed by means of a hybrid constructive method to, then, be submitted to the standard local search procedure of the algorithm. A pattern $p$ is selected in round-robin fashion and it is given as input to the hybrid constructive method. This method starts by inserting all elements of $p$ in the partial solution and completes this solution using the original constructive method. As a result, the solution built by this hybrid method will contain the elements of $p$, leading the exploration of the search space for better solutions as proposed. It is noteworthy to highlight that this hybridization process does not modify the local search phase in any aspect. In \cite{Ribeiro2004,Ribeiro2006}, different versions of DM-GRASP varying $d$, $sup_{min}$, type of patterns (e.g., maximal or non-maximal patterns), and the number of mined patterns, were tested. In general, the best results were achieved by a version of DM-GRASP based on maximal frequent itemsets, which is the mining strategy adopted in this work.
	
	Unlike DM-GRASP, another data mining strategy, named as MDM-GRASP (Multi Data Mining GRASP), has its mining process executed whenever the elite set becomes stable -- which refers to a number of iterations without any changes in the elite set. Firstly approached in \cite{Plastino2011}, the basic idea of MDM-GRASP is to execute the mining process: (a) as soon as the elite set becomes stable and (b) whenever the elite set has been changed and again becomes stable. Thereby, in practical terms, the quality of the elite set is progressively enhanced, yielding, at each mining process, more refined patterns.
	
	Initially, DM-GRASP and MDM-GRASP were successfully applied to optimization problems that represent solutions as sets of elements \citep{Santos2005, Santos2006, Plastino2011, Barbalho2013, Martins2018}. The mining of itemsets for this category of problems was straightforward since itemsets are sets of solution elements. However, this technique cannot be directly employed when the ordering of the elements is relevant in the solution, i.e., when the solution is a permutation.
	
	For solving problems characterized by permutations using data mining techniques, \cite{Guerine2016} proposed a strategy to deal with permutations tackling a Traveling Salesman Problem~(TSP) variant. Instead of the intuitive mining of vertices, which are the solution elements, this technique considered the mining of arcs -- which naturally holds the ordering of the vertices. Each pair of vertices ($i,j$), an arc of a solution, is mapped to an identifier that represents it ($a_{i\rightarrow j}$). This mapping is performed to all solutions from the elite set, i.e., transforming the solutions into sets of (arc) identifiers, which can then be mined by an FIM technique.	
	
	\subsection{Hybridization of GILS-RVND with Data Mining} \label{subsec:hybridization-gils-rvnd-mdm}
	
	This subsection describes the hybrid heuristic with data mining proposed in this paper, called MDM-GILS-RVND, as it is based on the MDM-GRASP approach. It must be pointed out that an algorithm based on DM-GRASP approach and GILS-RVND, named DM-GILS-RVND, has also been developed in this work. Although its results over GILS-RVND were promising, the MDM-GILS-RVND achieved better performance compared to GILS-RVND and DM-GILS-RVND, both in terms of solution quality and time. Therefore, we omitted the results of DM-GILS-RVND in this work.
	
	In a classic MDM-GRASP algorithm, the data mining procedure is carried out several times throughout the algorithm execution based on the stability of the elite set. However, this idea cannot be directly applied to GILS-RVND, since only ten multi-start iterations were defined for it in Silva et al. (2012), which is an insufficient number of iterations to let the elite set becomes stable. It is worth saying that the MDM-GRASP approach was applied to heuristics that required 200, 500, and 1000 iterations, respectively reported in \cite{Barbalho2013} \cite{Plastino2014} and \cite{Guerine2016}, showing promising results when compared with DM-GRASP strategy.
	
	Different from the classic MDM-GRASP, the proposed MDM-GILS-RVND, the mining procedure is executed in the following moments: (a) after the first half multi-start iterations and (b) whenever the elite set is updated. Therefore, MDM-GILS-RVND is an algorithm composed of two main phases, a phase for storing the best visited solutions in a set (elite set), and a phase for using patterns extracted from the elite set into initial solutions to guide the local search phase for better solutions. Each phase of MDM-GILS-RVND executes a half of the total number of multi-start iterations of GILS-RVND.
	
	In its first phase, which is identical to the original heuristic, the best visited solutions are stored in an elite set $D$ of size $d$. A solution $s$ is eligible to be included in $D$ either $s$ is better than the worst solution of $D$, or $D$ is not full. Also, $D$ does not allow identical solutions. Then, the second phase executes, at each iteration, a data mining process in order to generate a list of patterns $P$ when $D$ has been updated. A pattern $p \in P$ is then used to construct an initial solution by means of a hybrid constructive procedure. In this phase, solutions visited in the local search can also be included in $D$. After the end of the second phase, the best solution found is returned.
	
\begin{figure}[htbp]
		\centering
		\subfigure[Solution S1]{
			\label{fig:solution1}
			\includegraphics[scale=0.22]{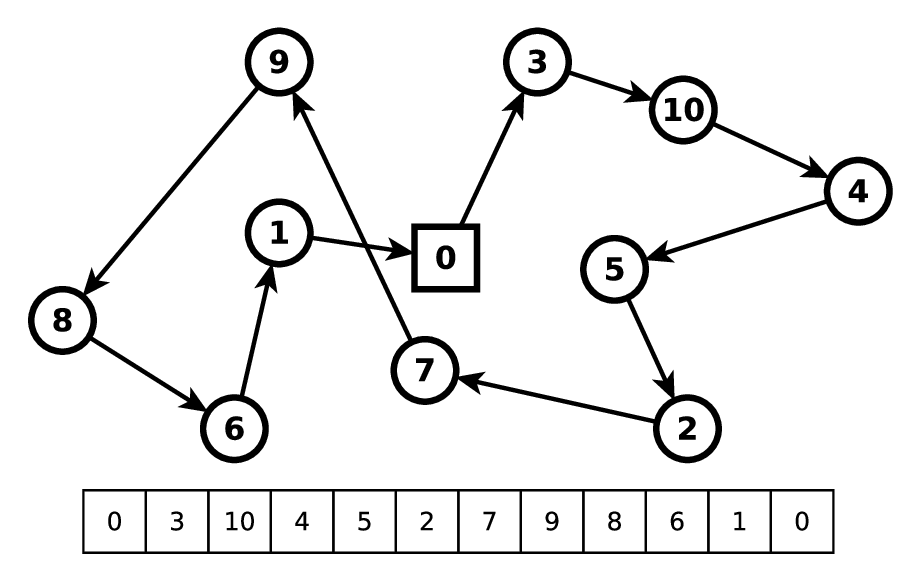}
		}
		\subfigure[Solution S2]{
			\label{fig:solution2}
			\includegraphics[scale=0.22]{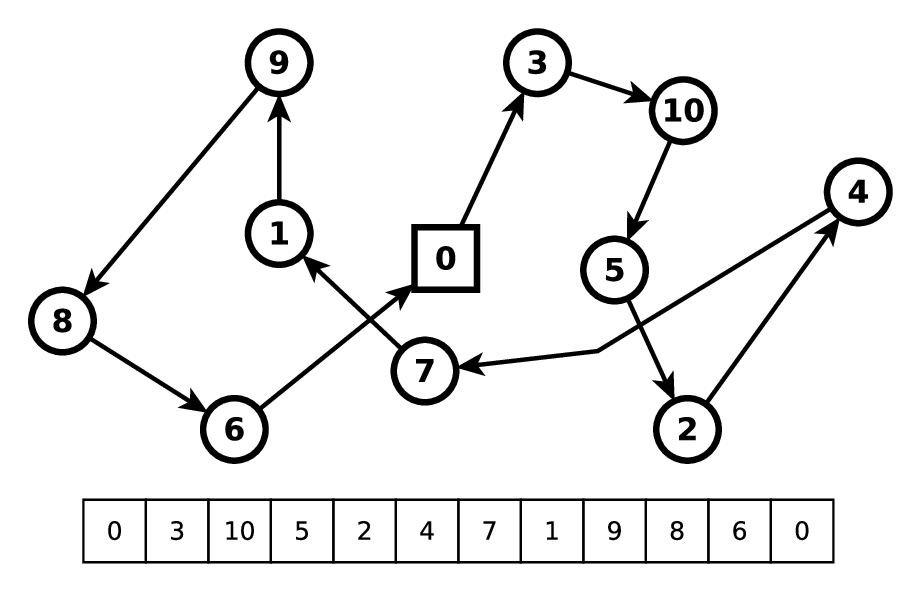}
		}
		\subfigure[Frequent arcs]{
			\label{fig:mined-arcs}
			\includegraphics[scale=0.22]{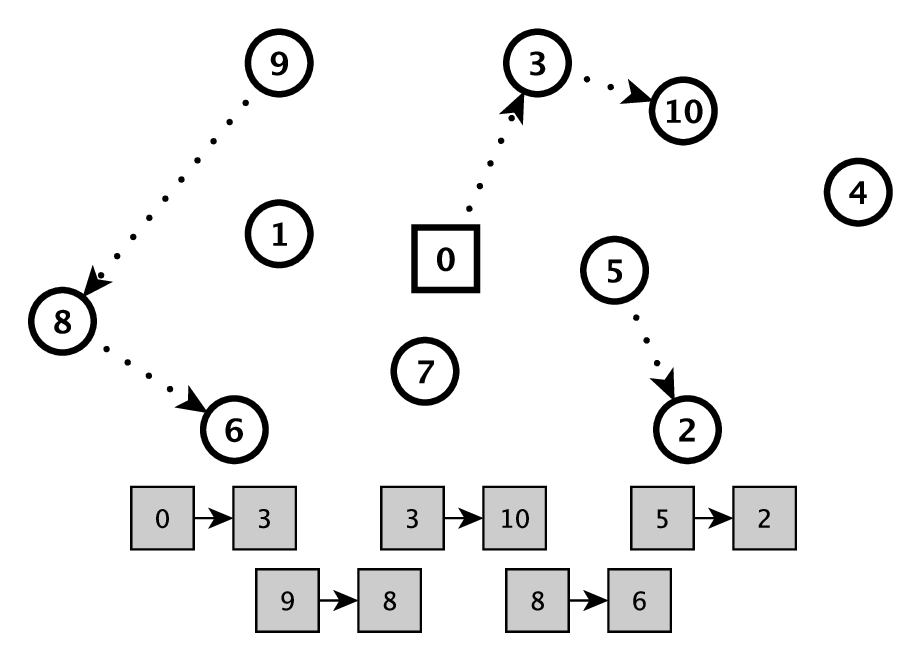}
		}
		\subfigure[A pattern-based solution]{
			\label{fig:generated-solution-with-frequent-arcs}
			\includegraphics[scale=0.22]{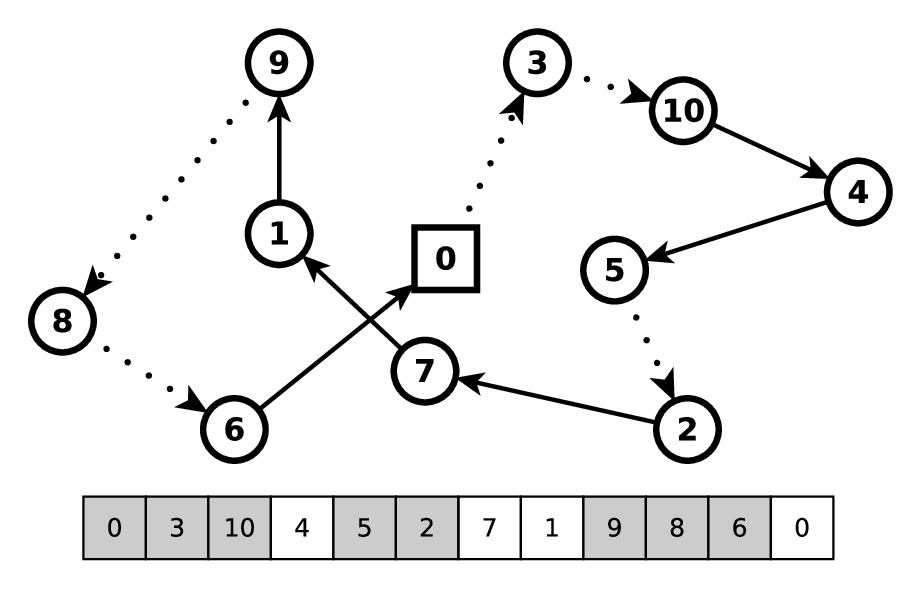}
		}
		\caption{Use of mined frequent arcs to generate initial solutions}
		\label{fig:mining-process}
	\end{figure}

	Fig.~\ref{fig:mining-process} wraps the main idea behind the hybridization applied to GILS-RVND in an illustrative scenario. Assume two solutions, S1 and S2, stored in an elite set of size two and $sup_{min}$ equal to 100\%. Since MLP solutions are permutations, the mapping approach developed in~\cite{Guerine2016} is adopted for the mining arcs. As a result, a pattern, i.e., a frequent set of (arc) identifiers, is extracted from the elite set. The mined pattern is depicted in dotted lines in Fig.~\ref{fig:mined-arcs}. Based on this pattern, a solution, shown in Fig.~\ref{fig:generated-solution-with-frequent-arcs}, is then generated by the hybrid constructive procedure, which is later explained. Note that all consecutive arcs are inserted in sequence as per Fig.~\ref{fig:generated-solution-with-frequent-arcs}.

	The details of MDM-GILS-RVND are displayed in Algorithm \ref{alg:mdm-gils-rvnd}. In the beginning, the cost of the best global solution $s^*$ is initialized (line 1). In the first phase of the strategy (lines 2-19), which comprises the first half of the iterations, the algorithm remains the same as GILS-RVND, except for the presence of the $UpdateEliteSet$($D$,~$d$,~$s$) function (line~8). This function inserts $s$, which is a solution returned by the RVND procedure, in $D$ if $s$ is unique in~$D$, and its cost value is better than the worst solution cost value in $D$. Next, a boolean flag $Updated$ is initialized and assigned to \textit{TRUE} (line~20) to enforce that the data mining process will be executed before the first multi-start iteration of the second phase  (line~23). In the second phase itself (lines 20-45), if $D$ (elite set) has been changed on the very last iteration, i.e., it was updated ($Updated$ is \textit{TRUE}), then, the mining process is performed again, and $Updated$ is set to \textit{FALSE} (lines~22-25). Indeed, the data mining process requires $D$, $d$, $sup_{min}$ and $MaxP$ as input parameters, where $MaxP$ is an integer threshold for the number of selected patterns (line~23). In our hybridization, the list of patterns $P$ (of size~$MaxP$) is composed of the largest patterns found by the FIM technique sorted in decreasing order by their size, i.e., the number of arcs. Then, the hybrid constructive procedure is called (line 26), which is detailed in the following paragraphs. In the local search loop (lines 29-41), new candidate solutions can be included in $D$ (line 31). If it happens, then $Updated$ is assigned to \textit{TRUE} (line 33), making the mining process be executed at the next multi-start iteration (line 23). Next, right after the local search phase, the best global solution $s^*$ may be updated (lines 42-44). When all multi-start iterations are executed, the algorithm stops, and $s^*$ is returned (line 46).
	
	In this work, the extraction of patterns is performed by the FPMax* algorithm\footnote{FPMax* is available at \url{http://fimi.uantwerpen.be/}.}, which is an efficient implementation for extracting maximal frequent itemsets~\citep{Grahne2003}. Although this implementation was presented in 2003, according to \cite{Yun2016}, FPMax* is still considered as a state-of-the-art algorithm along with MAFIA and LCM algorithms. In the context of our work, each solution (or permutation) of the elite set is transformed into a set of (arc) identifiers. Next, FPMax* performs the extraction of maximal patterns on these set-based solutions, where a mined pattern corresponds to a set of frequent arcs.

	{\linespread{0.9}
		\begin{algorithm}
			\caption{MDM-GILS-RVND$(I_{Max},I_{ILS}, R, D, d, Sup, MaxP)$}\label{alg:mdm-gils-rvnd}
			\begin{algorithmic}[1]
				\STATE $f(s^*) \leftarrow \infty;$
				\FOR{$i = 1, \dots,  I_{Max}/2$ }
				\STATE $s \leftarrow ConstructiveProcedure(\alpha \in R);$
				\STATE $s' \leftarrow s;$
				\STATE $iterILS \leftarrow 0;$
				\WHILE{$iterILS < I_{ILS}$}
				\STATE $s \leftarrow RVND(s);$
				\STATE $UpdateEliteSet(D, d, s);$
				\IF{$f(s) < f(s')$}
				\STATE $s' \leftarrow s;$
				\STATE $iterILS \leftarrow 0;$
				\ENDIF
				\STATE $s \leftarrow Perturb(s');$
				\STATE $iterILS \leftarrow iterILS + 1;$
				\ENDWHILE
				\IF{$ f(s') < f(s^*) $}
				\STATE $s^* \leftarrow s';$
				\ENDIF
				\ENDFOR
				\STATE $Updated \leftarrow$ \textit{TRUE};
				\FOR{$i = 1, \dots,  I_{Max}/2$ }
				\IF{$Updated$} 
				\STATE $P \leftarrow MinePatterns(D, d, Sup, MaxP);$
				\STATE $Updated \leftarrow$ \textit{FALSE};
				\ENDIF
				
				\STATE $s \leftarrow HybridConstructiveProc(\alpha \in R, p \in P);$
				\STATE $s' \leftarrow s;$
				\STATE $iterILS \leftarrow 0;$
				\WHILE{$iterILS < I_{ILS}$}
				\STATE $s \leftarrow RVND(s);$
				\STATE $UpdateEliteSet(D, d, s);$
				\IF{$D$~was~updated}
				\STATE $Updated \leftarrow$ \textit{TRUE};
				\ENDIF
				\IF{$f(s) < f(s')$}
				\STATE $s' \leftarrow s;$
				\STATE $iterILS \leftarrow 0;$
				\ENDIF
				\STATE $s \leftarrow Perturb(s');$
				\STATE $iterILS \leftarrow iterILS + 1;$
				\ENDWHILE
				\IF{$ f(s') < f(s^*) $}
				\STATE $s^* \leftarrow s';$
				\ENDIF
				\ENDFOR
				\STATE \textbf{return} $ s^*;$
			\end{algorithmic}
		\end{algorithm}
	}
	
	For generating an initial solution using a given pattern, the hybrid constructive procedure tries to insert segments of solutions in the partial solution when possible. We refer to a segment of solution as a sequence of consecutive arcs found in an extracted pattern. Considering the example shown in Fig.~\ref{fig:mining-process}, three segments were found and used to construct an initial solution. It is important to emphasize that during the construction process, only the first vertex of a segment is available to be chosen, so that when any first vertex is selected, its corresponding segment is included in the partial solution at once. 
	
	This hybrid constructive procedure is detailed in Algorithm 5. Firstly, the depot (vertex 0) is placed into the partial solution $s$ (line~1). Then, the Consecutive Arcs Lists (CAL) are generated using the pattern $p$ as input (line~2). Each list in CAL is a segment of solution found in $p$. Based on CAL, the Customers List (CL) of this algorithm is then created with those that do not exist in any list of CAL and those that are the very first customer of each list in CAL (line 3). Next, if the vertex 0 is the first vertex of a list in CAL, then this list is appended to $s$, and removed from CAL (lines 4-7). Then, the variable $r$ receives the last inserted vertex in $s$ (line 8). In the inner loop (lines 9-21), $r$ is used as reference in the sorting of CL (line 10), as already explained in Algorithm~\ref{alg:gils-rvnd-constructive}. Next, the RCL is filled with the ($\alpha$ x 100)\% best customers from CL (line~11). Then, a customer $c$ is randomly selected from RCL (line~12). The customer $c$ is checked in the first element of each CAL list and, if found, the entire segment is included in $s$ and removed from CAL (lines 13-15), otherwise, only $c$ is included in $s$ (lines 16-17). After this conditional block, $r$ is updated and $c$ is removed from CL (lines 19-20). When CL becomes empty, $s$ is returned by the algorithm (line 22)
	
	\begin{algorithm}
		\caption{HybridConstructiveProcedure$(\alpha, p)$}\label{alg:mdm-gils-rvnd-constructive}
		\begin{algorithmic}[1]
			\STATE $s \leftarrow \{0\};$
			\STATE $CAL \leftarrow GenerateCAL(p);$
			\STATE $CL \leftarrow GenerateCLFromCAL(CAL);$
			\IF{\{$\exists~cal \in CAL$ $|$ 0 is the first customer in $cal$\}}
			\STATE $s \leftarrow s \cup cal;$
			\STATE $CAL \leftarrow CAL - cal;$
			\ENDIF
			\STATE $r \leftarrow SelectLastCustomer(s);$
			\WHILE{$CL \neq \emptyset $}
			\STATE $CL \leftarrow SortCL(CL,r);$
			\STATE $RCL \leftarrow FillRCL(CL,\alpha);$
			\STATE $c \leftarrow SelectRandomClient(RCL);$
			\IF{\{$\exists~cal \in CAL$ $|$ $c$ is the first customer in $cal$\}}
			\STATE $s \leftarrow s \cup cal;$
			\STATE $CAL \leftarrow CAL - cal;$
			\ELSE
			\STATE $s \leftarrow s \cup \{c\};$
			\ENDIF
			\STATE $r \leftarrow SelectLastCustomer(s);$
			\STATE $CL \leftarrow CL - \{c\}$
			\ENDWHILE
			\STATE \textbf{return} $s$
		\end{algorithmic}
	\end{algorithm}
	
	\section{Computational Results}\label{sec:computational-experiments}

	The original GILS-RVND source-code was granted by the authors to develop our proposed MDM-GILS-RVND. This source-code was implemented in C++ and compiled with the g++~4.4.3, in single thread of a Intel\textsuperscript{\textregistered}Core\textsuperscript{TM} i7 3.40GHz with 16GB of RAM under GNU/Linux Ubuntu~14.04 (64-bits).
	
	All computational experiments reported in \cite{Silva2012} were fully re-executed using the same set of parameters, the same g++ compiler and the same seeds of pseudo-random numbers. For these experiments $I_{Max}$$=$$10$, $I_{ILS}$$=$$min\{100,n\}$ and $R$$=$$\{0.00,0.01,\dots,0.25\}$ are given as input parameters in both heuristics. Also, as indicated in \cite{Silva2012} and reproduced in our paper, each heuristic was run ten times for each instance, which used distinct seeds of pseudo-random numbers. Exclusively for MDM-GILS-RVND, $sup_{min}$$=$$70$\%, $d$$=$$10$ and $MaxP$$=$$5$ were defined. In fact, $sup_{min}$$=$$70\%$ was chosen based on the best trade-off between solution quality and computational time found in our experiments, different from the usual $sup_{min}$$=$$20\%$, adopted in~\cite{Santos2006,Santos2005,Plastino2014,Maia2016,Martins2018}, that presented impracticable averages of computational time due to the mining process. The elite set with size $d$$=$$10$ was based on promising results of MDM-GRASP \citep[see][]{Plastino2014, Maia2016, Guerine2016, Martins2018}. Finally, the definition of $MaxP=5$ considers that if the data mining procedure is executed only  once, then necessarily there will be five patterns in $P$, one for each of the five multi-start iterations of the second phase of MDM-GILS-RVND.
	
	Statistical tests were carried out on all instances results in order to verify whether the performance of both heuristics for each instance is, indeed, different and not merely at random. For each instance, a heuristic's sample stands for the ten solution cost values obtained by solving the corresponding instance. We adopted two paired one-tailed tests: the parametric Student's t-test, and the nonparametric Wilcoxon signed-rank test. If both samples follow a normal distribution tested by the Shapiro-Wilk test, then the Student's t-test is applied. Otherwise, the Wilcoxon's test is chosen. According to \cite{Siegel1988}, these tests are commonly used to compare two paired (dependent) samples, where their null hypothesis denotes that both heuristics have equal performance, and the alternative hypothesis means that one heuristic performs better than the another one. In our tests, the null hypothesis is rejected if the computed p-value is lower than 5\%.
	
	For Subsections \ref{subsec:experiments-circuit} and \ref{subsec:experiments-path}, tables are used to compare the results obtained by both heuristics. In a table, the first column represents the tested instance, and the following one stands for its best known solution (BKS) in the literature. For each heuristic, \textbf{Best Solution}, \textbf{Average Solution}, the percentage gap between \textbf{Average Solution} and \textbf{BKS} columns, and \textbf{Average Time} (in seconds) are shown. The next column reports the percentage gap between the \textbf{Average Time} columns of both heuristics. Values of the gap columns are obtained following $Gap(\%) = (100\times(\textit{MDMValue} - \textit{GILSValue})/\textit{GILSValue})$, where \textit{MDMValue} and \textit{GILSValue} mean, respectively, the obtained values by MDM-GILS-RVND and GILS-RVND. In the last column, we present the \textbf{p-value} obtained by the corresponding statistical test, which is underlined if there is statistical significance. Additionally, shown at the bottom of the table, averages for each gap column are exhibited at the \textbf{Average} row, and a counter of better results representing the number of best results of a heuristic over the other at the \textbf{Better} row, eligible only for columns of \textbf{Best Solution}, \textbf{Average Solution} and \textbf{Average Time}.

	\subsection{Experiments for Hamiltonian Circuit} \label{subsec:experiments-circuit}

	This subsection presents computational results for the MLP variant that considers Hamiltonian circuits as solutions, where two instance sets were tested. A set of 23 instances varying from 42 to 107 customers, originally selected in~\cite{Abeledo2010a, Abeledo2010b} from TSPLib, and another set composed of all 56 instances with 120 to 1379 customers from TSPLib. In the latter set, 18 instances have been already tested in~\cite{Bulhoes2018}, and the other 38 instances will be used for the first time in this work.
	
	Since the results on the 23-instances set selected in~\cite{Abeledo2010a, Abeledo2010b} were virtually identical in terms of solution quality, these results were not shown in a table. For best solutions, both heuristics achieved optimal solutions in all instances. Regarding average solutions, both MDM-GILS-RVND and GILS-RVND found the same results in 20 instances. For the remaining three instances, the original heuristic found slightly better results, but without statistical difference. In terms of computational time, MDM-GILS-RVND performed 5.48\% faster than GILS-RVND.
	
	Table \ref{tab:mdm-results-circuit-new} reports the results on the 56-instances set. For average solutions, the data mining proposal achieved 30 better results than GILS-RVND, while this heuristic outperformed MDM-GILS-RVND in 11 results. On the best solution aspect, MDM-GILS-RVND again achieved better results, since the original heuristic obtained 40 best results against 49 results from MDM-GILS-RVND. Beyond solution quality, MDM-GILS-RVND performed faster than GILS-RVND: the average running time was reduced in 23.36\%. There were seven results with statistical significance when MDM-GILS-RVND was better than GILS-RVND, whereas one result was statistically significant when GILS-RVND was better than MDM-GILS-RVND. Lastly, considering the BKS aspect, both heuristics achieved all solution cost values reported in~\cite{Bulhoes2018}.
	
	\begin{table}[htbp]
		\centering
		\caption{Results on the 56-instances set selected from TSPLib}
		{
			\resizebox{0.95\textwidth}{!}{
				\small
				\begin{tabular}{lrrrrrrrrrrrr}
					\hline
					&  &  \multicolumn{4}{c}{GILS-RVND} & &
					\multicolumn{4}{c}{MDM-GILS-RVND}  \\   \cline{3-6} \cline{8-11}
					\multirow{2}{*}{Instance} & 
					\multirow{2}{*}{\begin{minipage}{0.4in}\begin{center}BKS\end{center}\end{minipage}} & 
					
					\multirow{2}{*}{\begin{minipage}{0.4in}\begin{center}Best\\Solution\end{center}\end{minipage}} &
					\multirow{2}{*}{\begin{minipage}{0.4in}\begin{center}Average Solution\end{center}\end{minipage}} &
					\multirow{2}{*}{\begin{minipage}{0.25in}\begin{center}BKS\\Gap(\%)\end{center}\end{minipage}} &
					\multirow{2}{*}{\begin{minipage}{0.4in}\begin{center}Average Time(s)\end{center}\end{minipage}} & &
					
					\multirow{2}{*}{\begin{minipage}{0.4in}\begin{center}Best\\Solution\end{center}\end{minipage}} &
					\multirow{2}{*}{\begin{minipage}{0.4in}\begin{center}Average Solution\end{center}\end{minipage}} &
					\multirow{2}{*}{\begin{minipage}{0.25in}\begin{center}BKS\\Gap(\%)\end{center}\end{minipage}} &
					\multirow{2}{*}{\begin{minipage}{0.4in}\begin{center}Average Time(s)\end{center}\end{minipage}} & 
					\multirow{2}{*}{\begin{minipage}{0.35in}\begin{center}Time Gap(\%)\end{center}\end{minipage}} &
					\multirow{2}{*}{\begin{minipage}{0.35in}\begin{center}p-value\end{center}\end{minipage}} \\ \\
					\hline
 gr120&$^{\beta,\alpha}$363454&\textbf{363454}&363569.5&0.03&9.54&&\textbf{363454}&\textbf{363454.0}&0.00&\textbf{8.10}&-15.09&$^\omega$0.500\\
pr124&$^{\beta,\alpha}$3154346&\textbf{3154346}&\textbf{3154346.0}&0.00&5.39&&\textbf{3154346}&\textbf{3154346.0}&0.00&\textbf{5.15}&-4.45&\multicolumn{1}{c}{-}\\
bier127&$^{\beta,\alpha}$4545005&\textbf{4545005}&4546378.8&0.03&9.25&&\textbf{4545005}&\textbf{4545691.9}&0.02&\textbf{7.73}&-15.68&$^\omega$0.386\\
ch130&$^{\beta,\alpha}$349874&\textbf{349874}&\textbf{349891.7}&0.01&9.23&&\textbf{349874}&349903.5&0.01&\textbf{8.88}&-3.79&$^\omega$0.173\\
pr136&$^{\beta,\alpha}$6199268&\textbf{6199268}&\textbf{6199805.4}&0.01&17.30&&\textbf{6199268}&6200041.6&0.01&\textbf{14.82}&-14.34&$^\omega$0.605\\
gr137&$^{\beta,\alpha}$4061498&\textbf{4061498}&\textbf{4061498.0}&0.00&8.11&&\textbf{4061498}&\textbf{4061498.0}&0.00&\textbf{7.16}&-11.71&\multicolumn{1}{c}{-}\\
pr144&$^{\beta,\alpha}$3846137&\textbf{3846137}&\textbf{3846137.0}&0.00&9.11&&\textbf{3846137}&\textbf{3846137.0}&0.00&\textbf{8.80}&-3.40&\multicolumn{1}{c}{-}\\
ch150&$^{\beta,\alpha}$444424&\textbf{444424}&\textbf{444424.0}&0.00&13.06&&\textbf{444424}&\textbf{444424.0}&0.00&\textbf{10.67}&-18.30&\multicolumn{1}{c}{-}\\
kroA150&$^{\beta,\alpha}$1825769&\textbf{1825769}&\textbf{1825769.0}&0.00&19.84&&\textbf{1825769}&\textbf{1825769.0}&0.00&\textbf{15.68}&-20.97&\multicolumn{1}{c}{-}\\
kroB150&$^{\beta,\alpha}$1786546&\textbf{1786546}&\textbf{1786546.0}&0.00&16.27&&\textbf{1786546}&\textbf{1786546.0}&0.00&\textbf{14.49}&-10.94&\multicolumn{1}{c}{-}\\
pr152&$^{\beta,\alpha}$5064566&\textbf{5064566}&\textbf{5064566.0}&0.00&11.23&&\textbf{5064566}&\textbf{5064566.0}&0.00&\textbf{10.20}&-9.17&\multicolumn{1}{c}{-}\\
u159&$^{\beta,\alpha}$2972030&\textbf{2972030}&\textbf{2972204.2}&0.01&14.21&&\textbf{2972030}&\textbf{2972204.2}&0.01&\textbf{12.92}&-9.08&$^\omega$0.681\\
si175&$^{\beta,\alpha}$1808532&\textbf{1808532}&\textbf{1808532.0}&0.00&19.14&&\textbf{1808532}&\textbf{1808532.0}&0.00&\textbf{14.85}&-22.41&\multicolumn{1}{c}{-}\\
brg180&$^{\beta,\alpha}$174750&\textbf{174750}&\textbf{174750.0}&0.00&16.79&&\textbf{174750}&\textbf{174750.0}&0.00&\textbf{16.18}&-3.63&\multicolumn{1}{c}{-}\\
rat195&$^{\beta,\alpha}$218632&\textbf{218632}&218763.2&0.06&44.69&&\textbf{218632}&\textbf{218736.6}&0.05&\textbf{35.57}&-20.41&$^\tau$0.278\\
d198&$^{\beta}$1186049&\textbf{1186049}&\textbf{1186098.6}&0.00&38.28&&\textbf{1186049}&1186273.3&0.02&\textbf{31.95}&-16.54&$^\omega$0.715\\
kroA200&$^{\beta}$2672437&\textbf{2672437}&\textbf{2672444.2}&0.00&42.23&&\textbf{2672437}&\textbf{2672444.2}&0.00&\textbf{33.73}&-20.13&$^\omega$0.681\\
kroB200&$^{\beta}$2669515&\textbf{2669515}&\textbf{2674486.0}&0.19&42.00&&\textbf{2669515}&2675993.6&0.24&\textbf{36.17}&-13.88&$^\omega$0.277\\
gr202&$\psi$&\textbf{2909247}&2914644.2&\multicolumn{1}{c}{-}&35.95&&\textbf{2909247}&\textbf{2913368.4}&\multicolumn{1}{c}{-}&\textbf{29.68}&-17.44&$^\omega$0.337\\
ts225&$\psi$&\textbf{13240046}&\textbf{13240046.0}&\multicolumn{1}{c}{-}&\textbf{26.60}&&\textbf{13240046}&\textbf{13240046.0}&\multicolumn{1}{c}{-}&27.10&1.88&\multicolumn{1}{c}{-}\\
tsp225&$\psi$&\textbf{402783}&403080.2&\multicolumn{1}{c}{-}&53.89&&\textbf{402783}&\textbf{402933.3}&\multicolumn{1}{c}{-}&\textbf{43.43}&-19.41&$^\omega$0.091\\
pr226&$\psi$&\textbf{7196869}&\textbf{7196869.0}&\multicolumn{1}{c}{-}&34.29&&\textbf{7196869}&\textbf{7196869.0}&\multicolumn{1}{c}{-}&\textbf{28.85}&-15.86&\multicolumn{1}{c}{-}\\
gr229&$\psi$&\textbf{10725914}&\textbf{10729883.8}&\multicolumn{1}{c}{-}&53.66&&\textbf{10725914}&10731249.9&\multicolumn{1}{c}{-}&\textbf{41.12}&-23.37&$^\omega$0.140\\
gil262&$\psi$&\textbf{285060}&285527.1&\multicolumn{1}{c}{-}&96.12&&\textbf{285060}&\textbf{285312.6}&\multicolumn{1}{c}{-}&\textbf{74.72}&-22.26&$^\omega$0.138\\
pr264&$\psi$&\textbf{5471615}&\textbf{5471615.0}&\multicolumn{1}{c}{-}&47.02&&\textbf{5471615}&\textbf{5471615.0}&\multicolumn{1}{c}{-}&\textbf{38.68}&-17.74&\multicolumn{1}{c}{-}\\
a280&$\psi$&\textbf{346989}&347125.9&\multicolumn{1}{c}{-}&107.18&&\textbf{346989}&\textbf{347106.9}&\multicolumn{1}{c}{-}&\textbf{79.05}&-26.25&$^\omega$0.383\\
pr299&$\psi$&\textbf{6556628}&\textbf{6557983.4}&\multicolumn{1}{c}{-}&104.92&&\textbf{6556628}&6559030.8&\multicolumn{1}{c}{-}&\textbf{75.93}&-27.63&$^\omega$0.500\\
lin318&$\psi$&\textbf{5619810}&\textbf{5629995.9}&\multicolumn{1}{c}{-}&117.98&&\textbf{5619810}&5630590.5&\multicolumn{1}{c}{-}&\textbf{90.74}&-23.09&$^\omega$0.541\\
rd400&$\psi$&2768830&2776672.7&\multicolumn{1}{c}{-}&350.84&&\textbf{2762532}&\textbf{2775707.0}&\multicolumn{1}{c}{-}&\textbf{247.61}&-29.42&$^\tau$0.339\\
fl417&$\psi$&\textbf{1874242}&1874242.8&\multicolumn{1}{c}{-}&382.64&&\textbf{1874242}&\textbf{1874242.0}&\multicolumn{1}{c}{-}&\textbf{250.61}&-34.51&$^\omega$0.500\\
gr431&$\psi$&\textbf{21159702}&21239150.9&\multicolumn{1}{c}{-}&336.98&&21180562.0&\textbf{21214270.9}&\multicolumn{1}{c}{-}&\textbf{245.10}&-27.27&$^\omega$0.091\\
pr439&$\psi$&\textbf{17829541}&17887107.0&\multicolumn{1}{c}{-}&285.56&&\textbf{17829541}&\textbf{17868632.7}&\multicolumn{1}{c}{-}&\textbf{200.02}&-29.96&$^\tau$\underline{0.046}\\
pcb442&$\psi$&\textbf{10301705}&10323539.7&\multicolumn{1}{c}{-}&413.41&&\textbf{10301705}&\textbf{10321465.7}&\multicolumn{1}{c}{-}&\textbf{291.64}&-29.46&$^\omega$0.172\\
d493&$\psi$&6684190&6691057.1&\multicolumn{1}{c}{-}&608.47&&\textbf{6677458}&\textbf{6687268.2}&\multicolumn{1}{c}{-}&\textbf{390.16}&-35.88&$^\tau$0.096\\
att532&$\psi$&\textbf{5613010}&5632753.5&\multicolumn{1}{c}{-}&988.04&&5617783.0&\textbf{5628346.4}&\multicolumn{1}{c}{-}&\textbf{760.49}&-23.03&$^\tau$0.133\\
ali535&$\psi$&31870389&\textbf{31904676.6}&\multicolumn{1}{c}{-}&880.76&&\textbf{31860679}&31910477.9&\multicolumn{1}{c}{-}&\textbf{570.79}&-35.19&$^\tau$0.331\\
si535&$\psi$&\textbf{12247211}&\textbf{12250679.7}&\multicolumn{1}{c}{-}&498.76&&12248066.0&12251841.0&\multicolumn{1}{c}{-}&\textbf{319.01}&-36.04&$^\tau$0.182\\
pa561&$\psi$&\textbf{658870}&\textbf{661211.6}&\multicolumn{1}{c}{-}&1155.32&&660590.0&661790.6&\multicolumn{1}{c}{-}&\textbf{873.20}&-24.42&$^\tau$0.167\\
u574&$\psi$&9314596&9344178.4&\multicolumn{1}{c}{-}&1234.19&&\textbf{9308820}&\textbf{9333295.3}&\multicolumn{1}{c}{-}&\textbf{854.63}&-30.75&$^\omega$0.203\\
rat575&$\psi$&1848869&1859221.1&\multicolumn{1}{c}{-}&1739.46&&\textbf{1847272}&\textbf{1856335.1}&\multicolumn{1}{c}{-}&\textbf{1234.05}&-29.06&$^\tau$0.089\\
p654&$\psi$&\textbf{7827273}&\textbf{7827639.2}&\multicolumn{1}{c}{-}&1755.28&&\textbf{7827273}&7827867.8&\multicolumn{1}{c}{-}&\textbf{1239.21}&-29.40&$^\tau$\underline{0.008}\\
d657&$\psi$&14159477&14220133.3&\multicolumn{1}{c}{-}&2615.66&&\textbf{14112540}&\textbf{14195797.6}&\multicolumn{1}{c}{-}&\textbf{1779.16}&-31.98&$^\tau$\underline{0.039}\\
gr666&$\psi$&63571693&63731966.5&\multicolumn{1}{c}{-}&2296.23&&\textbf{63500984}&\textbf{63612943.5}&\multicolumn{1}{c}{-}&\textbf{1609.15}&-29.92&$^\tau$\underline{0.021}\\
u724&$\psi$&13506660&13558605.3&\multicolumn{1}{c}{-}&4651.76&&\textbf{13504408}&\textbf{13537514.7}&\multicolumn{1}{c}{-}&\textbf{3132.70}&-32.66&$^\tau$\underline{0.031}\\
rat783&$\psi$&3282794&3296069.6&\multicolumn{1}{c}{-}&7044.52&&\textbf{3275858}&\textbf{3293606.1}&\multicolumn{1}{c}{-}&\textbf{4475.85}&-36.46&$^\tau$0.269\\
dsj1000&$\psi$&7646018508&7685887300.0&\multicolumn{1}{c}{-}&18068.70&&\textbf{7642715113}&\textbf{7664531851.0}&\multicolumn{1}{c}{-}&\textbf{12233.54}&-32.29&$^\omega$\underline{0.037}\\
dsj1000ceil&$\psi$&7646519008&7683329486.0&\multicolumn{1}{c}{-}&18543.76&&\textbf{7646395679}&\textbf{7676973751.0}&\multicolumn{1}{c}{-}&\textbf{11929.94}&-35.67&$^\tau$0.225\\
pr1002&$\psi$&115550770&116178260.2&\multicolumn{1}{c}{-}&11963.29&&\textbf{115420846}&\textbf{115874237.0}&\multicolumn{1}{c}{-}&\textbf{8029.58}&-32.88&$^\tau$0.093\\
si1032&$\psi$&\textbf{46896355}&46897662.4&\multicolumn{1}{c}{-}&2402.72&&\textbf{46896355}&\textbf{46896783.6}&\multicolumn{1}{c}{-}&\textbf{1926.99}&-19.80&$^\omega$0.172\\
u1060&$\psi$&\textbf{102508056}&102759766.0&\multicolumn{1}{c}{-}&15680.50&&102539819.0&\textbf{102759493.6}&\multicolumn{1}{c}{-}&\textbf{9910.02}&-36.80&$^\tau$0.499\\
vm1084&$\psi$&94760440&95053081.2&\multicolumn{1}{c}{-}&13894.43&&\textbf{94670122}&\textbf{94960603.3}&\multicolumn{1}{c}{-}&\textbf{9468.85}&-31.85&$^\tau$0.123\\
pcb1173&$\psi$&30926325&31032128.8&\multicolumn{1}{c}{-}&20508.89&&\textbf{30890385}&\textbf{30957008.7}&\multicolumn{1}{c}{-}&\textbf{14037.76}&-31.55&$^\tau$\underline{0.009}\\
d1291&$\psi$&\textbf{29383346}&\textbf{29477239.4}&\multicolumn{1}{c}{-}&12171.21&&29389729.0&29515210.4&\multicolumn{1}{c}{-}&\textbf{8072.84}&-33.67&$^\tau$0.147\\
rl1304&$\psi$&144886001&145596878.7&\multicolumn{1}{c}{-}&18617.53&&\textbf{144592447}&\textbf{145398549.2}&\multicolumn{1}{c}{-}&\textbf{12407.04}&-33.36&$^\tau$\underline{0.019}\\
rl1323&$\psi$&\textbf{155697857}&156360364.3&\multicolumn{1}{c}{-}&22758.06&&155719283.0&\textbf{156273365.5}&\multicolumn{1}{c}{-}&\textbf{15115.63}&-33.58&$^\tau$0.306\\
nrw1379&$\psi$&35360407&35519379.7&\multicolumn{1}{c}{-}&49624.72&&\textbf{35291795}&\textbf{35456093.0}&\multicolumn{1}{c}{-}&\textbf{32038.65}&-35.44&$^\tau$0.064\\
\hline
Average &&&&0.02&&&&&0.02&&-23.36 \\
Better &&7&11&&1&&16&30&&55&
\\ 
\hline
 \multicolumn{6}{l}{Legend:}\\
\multicolumn{6}{l}{$\beta$: cost values from \cite{Bulhoes2018}}\\
\multicolumn{6}{l}{$\alpha$: optimality is proven in~\citep{Bulhoes2018}}\\
\multicolumn{6}{l}{$\psi$: instances used for the first time in this work}\\
\multicolumn{6}{l}{$^\omega$: Wilcoxon's test applied}\\
\multicolumn{6}{l}{$^\tau$: Student's t-test applied}
				\end{tabular}
			}
		}
		\label{tab:mdm-results-circuit-new}
	\end{table}
	
	\subsection{Experiments for Hamiltonian Path} \label{subsec:experiments-path}
	
	For the MLP variant which considers Hamiltonian paths as solutions, 150 instances were tested and divided into a set of 10 instances varying between 70 and 532 customers selected from TSPLib, introduced in~\cite{Salehipour2011}, and seven groups of 20 generated instances each with dimensions of 10, 20, 50, 100, 200, 500, and 1000 customers by the authors of~\cite{Salehipour2011}.
	
	Computational results for the 10-instances set and the sets of 10, 20, 50, 100 and 200 customers are briefly discussed in the following paragraphs as both heuristics obtained very similar performance.
	
	In experiments for the 10-instances set, both heuristics reached the same best solutions, where nine of them are, indeed, BKSs. For average solutions, these heuristics presented very similar results, reflected by average BKS gap values of 0.08\% and 0.09\%, respectively, for MDM-GILS-RVND and GILS-RVND. No difference between the heuristics' results was statistically significant. In contrast to solution quality results, a computational time reduction of 14.47\% was obtained by MDM-GILS-RVND in the comparison between both heuristics.
	
	The results from sets of 10, 20, and 50 customers were entirely identical in both heuristics, where, for all instances, each average solution reached in the experiments was equal to the optimal solution. Moreover, each execution in all three sets was carried out, on average, in less than one second by both heuristics.
	
	Considering the 100-customers and 200-customers sets, both heuristics also obtained the same best solutions, which all of them are BKSs. In terms of average solution for the 100-customers set, MDM-GILS-RVND and GILS-RVND presented small average BKS gap values: 0.006\% and 0.002\%, respectively, without statistically significant results. For the 200-customers set, both heuristics achieved average BKS gap values of 0.03\%, without statistically significant results. Comparisons over these two sets showed a significant reduction of computational time. On average, MDM-GILS-RVND performed 10.41\% and 21.18\% faster than GILS-RVND, respectively, for the 100-customers and 200-customers sets.
	
	Computational results on the 500-customers set are reported in Table~\ref{tab:mdm-quality-s500}. Regarding best solution, MDM-GILS-RVND obtained 15 better results compared to five results from GILS-RVND, while for average solution, the data mining heuristic attained 18 better results compared to only two results reached by the original heuristic. A smaller gap between average solution and BKS was achieved by MDM-GILS-RVND with an average of 0.41\%, while GILS-RVND under the same conditions got 0.51\%. In running time, MDM-GILS-RVND performed better than the original heuristic in all cases, reducing in 27.74\% the average of computational time. Five instances, where MDM-GILS-RVND was better than GILS-RVND, were statistically significant for Student's t-test.

	\begin{table}[h]
		\centering
		\caption{Results for the 500-customers set generated in \cite{Salehipour2011}}
		\resizebox{\textwidth}{!}
		{\small
			\begin{tabular}{lrrrrrrrrrrrr}
				\hline
				&  &  \multicolumn{4}{c}{GILS-RVND} & &
				\multicolumn{4}{c}{MDM-GILS-RVND}  \\   \cline{3-6} \cline{8-11}
				\multirow{2}{*}{Instance} & 
				\multirow{2}{*}{\begin{minipage}{0.4in}\begin{center}BKS\end{center}\end{minipage}} & 
				
				\multirow{2}{*}{\begin{minipage}{0.4in}\begin{center}Best\\Solution\end{center}\end{minipage}} &
				\multirow{2}{*}{\begin{minipage}{0.4in}\begin{center}Average Solution\end{center}\end{minipage}} &
				\multirow{2}{*}{\begin{minipage}{0.25in}\begin{center}BKS\\Gap(\%)\end{center}\end{minipage}} &
				\multirow{2}{*}{\begin{minipage}{0.4in}\begin{center}Average Time(s)\end{center}\end{minipage}} & &
				
				\multirow{2}{*}{\begin{minipage}{0.4in}\begin{center}Best\\Solution\end{center}\end{minipage}} &
				\multirow{2}{*}{\begin{minipage}{0.4in}\begin{center}Average Solution\end{center}\end{minipage}} &
				\multirow{2}{*}{\begin{minipage}{0.25in}\begin{center}BKS\\Gap(\%)\end{center}\end{minipage}} &
				\multirow{2}{*}{\begin{minipage}{0.4in}\begin{center}Average Time(s)\end{center}\end{minipage}} & 
				\multirow{2}{*}{\begin{minipage}{0.35in}\begin{center}Time Gap(\%)\end{center}\end{minipage}} &
				\multirow{2}{*}{\begin{minipage}{0.35in}\begin{center}p-value\end{center}\end{minipage}} \\ \\
				\hline
				TRP-S500-R1&$^\star$1841386&\textbf{1841386}&1856018.7&0.79&830.85&&\textbf{1841386}&\textbf{1850046.4}&0.47&\textbf{623.37}&-24.97&$^\tau$\underline{0.037}\\
				TRP-S500-R2&$^\dagger$1815664&\textbf{1816568}&1823196.9&0.41&724.17&&1817057&\textbf{1822540.1}&0.38&\textbf{485.98}&-32.89&$^\tau$0.223\\
				TRP-S500-R3&$^\dagger$1826855&1833044&1839254.2&0.68&761.86&&\textbf{1827550}&\textbf{1837315.8}&0.57&\textbf{571.46}&-24.99&$^\tau$0.204\\
				TRP-S500-R4&$^\dagger$1804894&1809266&1815876.4&0.61&810.63&&\textbf{1804005}&\textbf{1813295.1}&0.47&\textbf{579.34}&-28.53&$^\tau$0.162\\
				TRP-S500-R5&$^\dagger$1821250&1823975&1834031.7&0.70&734.32&&\textbf{1823135}&\textbf{1832073.0}&0.59&\textbf{512.22}&-30.25&$^\tau$0.107\\
				TRP-S500-R6&$^\dagger$1782731&1786620&1790912.4&0.46&796.28&&\textbf{1784189}&\textbf{1789923.1}&0.40&\textbf{612.53}&-23.08&$^\tau$0.214\\
				TRP-S500-R7&$^\star$1847999&1847999&1857926.6&0.54&781.01&&\textbf{1846753}&\textbf{1853596.3}&0.30&\textbf{620.16}&-20.60&$^\tau$\underline{0.001}\\
				TRP-S500-R8&$^\dagger$1819636&1820846&1829257.3&0.53&769.33&&\textbf{1820421}&\textbf{1828768.2}&0.50&\textbf{574.23}&-25.36&$^\tau$0.413\\
				TRP-S500-R9&$^\star$1733819&1733819&1737024.9&0.18&693.82&&\textbf{1731594}&\textbf{1736149.5}&0.13&\textbf{530.72}&-23.51&$^\tau$0.277\\
				TRP-S500-R10&$^\dagger$1761174&1762741&1767366.3&0.35&784.64&&\textbf{1761824}&\textbf{1766078.2}&0.28&\textbf{564.93}&-28.00&$^\omega$0.152\\
				TRP-S500-R11&$^\star$1797881&\textbf{1797881}&\textbf{1801467.9}&0.20&741.50&&\textbf{1797881}&1802827.0&0.28&\textbf{513.56}&-30.74&$^\tau$0.201\\
				TRP-S500-R12&$^\star$1774452&\textbf{1774452}&1783847.1&0.53&766.23&&\textbf{1774452}&\textbf{1783638.0}&0.52&\textbf{523.90}&-31.63&$^\tau$0.451\\
				TRP-S500-R13&$^\dagger$1863905&1873699&1878049.4&0.76&797.54&&\textbf{1867156}&\textbf{1875908.3}&0.64&\textbf{540.20}&-32.27&$^\tau$0.095\\
				TRP-S500-R14&$^\star$1799171&1799171&1805732.9&0.36&835.86&&\textbf{1796425}&\textbf{1802431.8}&0.18&\textbf{565.17}&-32.38&$^\tau$\underline{0.046}\\
				TRP-S500-R15&$^\dagger$1785263&1791145&1797532.9&0.69&800.69&&\textbf{1785155}&\textbf{1793916.3}&0.48&\textbf{556.92}&-30.44&$^\tau$0.101\\
				TRP-S500-R16&$^\dagger$1804392&1810188&\textbf{1816484.0}&0.67&761.93&&\textbf{1808775}&1817049.6&0.70&\textbf{555.91}&-27.04&$^\tau$0.407\\
				TRP-S500-R17&$^\star$1825748&1825748&1834443.2&0.48&738.57&&\textbf{1821971}&\textbf{1830454.8}&0.26&\textbf{548.44}&-25.74&$^\tau$\underline{0.019}\\
				TRP-S500-R18&$^\dagger$1825615&\textbf{1826263}&1833323.7&0.42&780.31&&\textbf{1826263}&\textbf{1831295.2}&0.31&\textbf{581.67}&-25.46&$^\tau$\underline{0.039}\\
				TRP-S500-R19&$^\dagger$1776855&1779248&1782763.9&0.33&773.39&&\textbf{1775023}&\textbf{1781604.6}&0.27&\textbf{559.35}&-27.68&$^\tau$0.239\\
				TRP-S500-R20&$^\star$1820813&1820813&1830483.3&0.53&726.50&&\textbf{1820168}&\textbf{1830222.6}&0.52&\textbf{514.63}&-29.16&$^\tau$0.423\\
				\hline
				Average&&&&0.51&&&&&0.41&&-27.74\\
				Better&&1&2&&0&&15&18&&20 \\ 
				\hline
				Legend:\\
				\multicolumn{4}{l}{$^\dagger$ cost values from \cite{Rios2016}}\\
				\multicolumn{4}{l}{$^\star$ cost values from \cite{Silva2012}}\\
				\multicolumn{10}{l}{$^\omega$ Wilcoxon's test applied}\\
				\multicolumn{10}{l}{$^\tau$ Student's t-test applied}
		\end{tabular}}
		
		\label{tab:mdm-quality-s500}
	\end{table}
	
	Table~\ref{tab:mdm-quality-s1000} presents results on the 1000-customers set, the hardest instance set of this MLP variant. Considering best solutions, the MDM heuristic improved 19 out of 20 results of GILS-RVND, whereas the average behavior of MDM-GILS-RVND (average solution aspect) was superior in all 20 instances, also presenting a smaller gap to BKS when compared to GILS-RVND results. As expected, MDM-GILS-RVND again attained the best performance of computational time over GILS-RVND, requiring, in average, 32.78\% less running time. Regarding statistical significance tests, ten instances, where MDM-GILS-RVND was better than GILS-RVND, were statistically significant for the Student's t-test.
	
	\begin{table}
		\centering
		\caption{Results on the 1000-customers set generated in \cite{Salehipour2011}}
		\resizebox{\textwidth}{!}
		{\small
			\begin{tabular}{lrrrrrrrrrrrr}
				\hline
				&  &  \multicolumn{4}{c}{GILS-RVND} & &
				\multicolumn{4}{c}{MDM-GILS-RVND}  \\   \cline{3-6} \cline{8-11}
				\multirow{2}{*}{Instance} & 
				\multirow{2}{*}{\begin{minipage}{0.4in}\begin{center}BKS\end{center}\end{minipage}} & 
				
				\multirow{2}{*}{\begin{minipage}{0.4in}\begin{center}Best\\Solution\end{center}\end{minipage}} &
				\multirow{2}{*}{\begin{minipage}{0.4in}\begin{center}Average Solution\end{center}\end{minipage}} &
				\multirow{2}{*}{\begin{minipage}{0.25in}\begin{center}BKS\\Gap(\%)\end{center}\end{minipage}} &
				\multirow{2}{*}{\begin{minipage}{0.4in}\begin{center}Average Time(s)\end{center}\end{minipage}} & &
				
				\multirow{2}{*}{\begin{minipage}{0.4in}\begin{center}Best\\Solution\end{center}\end{minipage}} &
				\multirow{2}{*}{\begin{minipage}{0.4in}\begin{center}Average Solution\end{center}\end{minipage}} &
				\multirow{2}{*}{\begin{minipage}{0.25in}\begin{center}BKS\\Gap(\%)\end{center}\end{minipage}} &
				\multirow{2}{*}{\begin{minipage}{0.4in}\begin{center}Average Time(s)\end{center}\end{minipage}} & 
				\multirow{2}{*}{\begin{minipage}{0.35in}\begin{center}Time Gap(\%)\end{center}\end{minipage}} &
				\multirow{2}{*}{\begin{minipage}{0.35in}\begin{center}p-value\end{center}\end{minipage}} \\ \\
				\hline
				TRP-S1000-R1&$^\star$5107395&5107395&5133698.3&0.52&19889.15&&\textbf{5097334}&\textbf{5117118.1}&0.19&\textbf{13747.69}&-30.88&$^\tau$\underline{0.043}\\
				TRP-S1000-R2&$^\star$5106161&5106161&5127449.4&0.42&19218.18&&\textbf{5083772}&\textbf{5106199.8}&$<$0.01&\textbf{13626.35}&-29.10&$^\tau$\underline{0.003}\\
				TRP-S1000-R3&$^\star$5096977&5096977&5113302.9&0.32&18798.18&&\textbf{5087014}&\textbf{5108822.4}&0.23&\textbf{11772.82}&-37.37&$^\tau$0.267\\
				TRP-S1000-R4&$^\dagger$5112465&5118006&5141392.6&0.57&18493.11&&\textbf{5110766}&\textbf{5137965.2}&0.50&\textbf{12985.91}&-29.78&$^\tau$0.245\\
				TRP-S1000-R5&$^\dagger$5097991&5103894&5122660.7&0.48&19143.87&&\textbf{5088230}&\textbf{5112437.5}&0.28&\textbf{13663.97}&-28.62&$^\tau$\underline{0.028}\\
				TRP-S1000-R6&$^\dagger$5109946&5115816&5143087.1&0.65&19143.87&&\textbf{5098630}&\textbf{5124817.1}&0.29&\textbf{13663.97}&-28.62&$^\tau$\underline{0.048}\\
				TRP-S1000-R7&$^\dagger$4995703&5021383&5032722.0&0.74&17681.02&&\textbf{4982461}&\textbf{5019363.8}&0.47&\textbf{11873.02}&-32.85&$^\tau$\underline{0.048}\\
				TRP-S1000-R8&$^\star$5109325&5109325&5132722.6&0.46&18065.24&&\textbf{5104125}&\textbf{5127143.8}&0.35&\textbf{12152.55}&-32.73&$^\tau$0.109\\
				TRP-S1000-R9&$^\dagger$5046566&5052599&5073245.3&0.53&17979.62&&\textbf{5044687}&\textbf{5069371.7}&0.45&\textbf{11744.68}&-34.68&$^\tau$0.294\\
				TRP-S1000-R10&$^\dagger$5060019&5078191&5093592.6&0.66&17596.33&&\textbf{5062172}&\textbf{5079835.2}&0.39&\textbf{12641.17}&-28.16&$^\tau$\underline{0.029}\\
				TRP-S1000-R11&$^\dagger$5031455&\textbf{5041913}&5066161.5&0.69&18307.69&&5051685&\textbf{5061519.8}&0.60&\textbf{11665.22}&-36.28&$^\tau$0.137\\
				TRP-S1000-R12&$^\star$5029792&5029792&5051235.2&0.43&19149.54&&\textbf{5023000}&\textbf{5038120.8}&0.17&\textbf{12801.08}&-33.15&$^\tau$\underline{0.042}\\
				TRP-S1000-R13&$^\star$5102520&5102520&5131437.5&0.57&19604.19&&\textbf{5085356}&\textbf{5114870.4}&0.24&\textbf{13484.77}&-31.21&$^\tau$\underline{0.033}\\
				TRP-S1000-R14&$^\dagger$5092861&5099433&5118980.6&0.51&18974.58&&\textbf{5087794}&\textbf{5111368.4}&0.36&\textbf{12877.75}&-32.13&$^\tau$0.130\\
				TRP-S1000-R15&$^\dagger$5131013&5142470&5174493.2&0.85&18889.53&&\textbf{5134114}&\textbf{5165423.3}&0.67&\textbf{12423.75}&-34.23&$^\tau$0.117\\
				TRP-S1000-R16&$^\dagger$5064094&5073972&5090280.5&0.52&18206.27&&\textbf{5047856}&\textbf{5088822.7}&0.49&\textbf{11437.22}&-37.18&$^\tau$0.438\\
				TRP-S1000-R17&$^\dagger$5052283&5071485&5084450.4&0.64&18571.62&&\textbf{5063904}&\textbf{5078517.9}&0.52&\textbf{11728.89}&-36.85&$^\tau$\underline{0.037}\\
				TRP-S1000-R18&$^\dagger$5005789&5017589&5037094.0&0.63&19745.37&&\textbf{4998254}&\textbf{5033070.8}&0.55&\textbf{12914.31}&-34.60&$^\tau$0.233\\
				TRP-S1000-R19&$^\dagger$5064873&5076800&5097167.6&0.64&19790.69&&\textbf{5065623}&\textbf{5086618.9}&0.43&\textbf{12793.33}&-35.36&$^\tau$\underline{0.036}\\
				TRP-S1000-R20&$^\star$4977262&4977262&5002920.6&0.52&18715.65&&\textbf{4976158}&\textbf{5000650.3}&0.47&\textbf{13099.58}&-30.01&$^\tau$0.357\\
				\hline
				Average &&&&0.57&&&&&0.38&&-32.78    \\
				Better&&1&0&&0&&19&20&&20\\ \hline
				Legend:\\
				\multicolumn{10}{l}{$^\dagger$ cost values from \cite{Rios2016}}\\
				\multicolumn{10}{l}{$^\star$ cost values from \cite{Silva2012}}\\
				\multicolumn{10}{l}{$^\omega$ Wilcoxon's test applied}\\
				\multicolumn{10}{l}{$^\tau$ Student's t-test applied}
			\end{tabular}
		}
		
		\label{tab:mdm-quality-s1000}
	\end{table}

	\subsection{Complementary Analyses} \label{subsec:complementary-analysis}
	
	This section presents three complementary assessments that consider different viewpoints of analyses to better understand the performances of GILS-RVND and MDM-GILS-RVND. The first one provides examples of the impact of using patterns into initial solutions and local search. The following analysis shows experiments of time convergence of GILS-RVND and MDM-GILS-RVND to targets (cost values of solutions). The last analysis presents comparisons of the evaluated heuristics using computational time as the stopping criterion.
	
	\subsubsection{Impact of the Usage of Mined Patterns} \label{subsubsec:impact-patterns}
	
	In this subsection, two executions for the TRP-S500-R17 instance, with two distinct seeds of pseudo-random numbers, were used to exemplify the impact of using mined patterns into initial solutions and the solution obtained after the local search phase. For each example, two figures show the results of each heuristic phase: the constructive phase and the local search phase, where the multi-start iterations were arranged at the abscissa axis and cost values of solutions at the ordinate axis.
	
\begin{figure}[htbp]
		\centering
		\subfigure[Constructive phase]{
			\label{fig:constructive-r17}
			\includegraphics[scale=0.57]{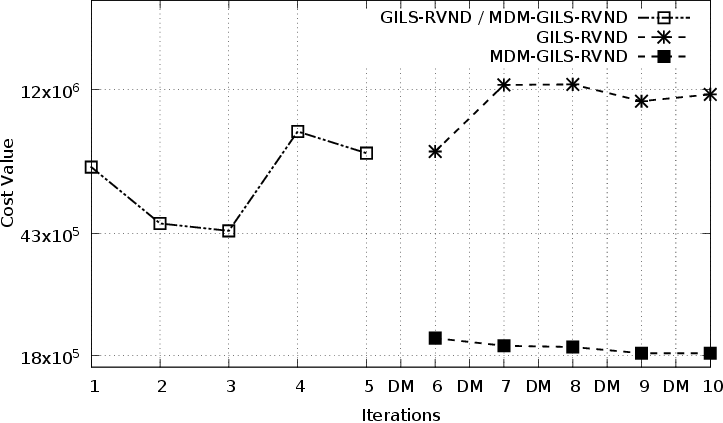}
		}
		\subfigure[Local search phase]{
			\label{fig:local-search-r17}
			\includegraphics[scale=0.57]{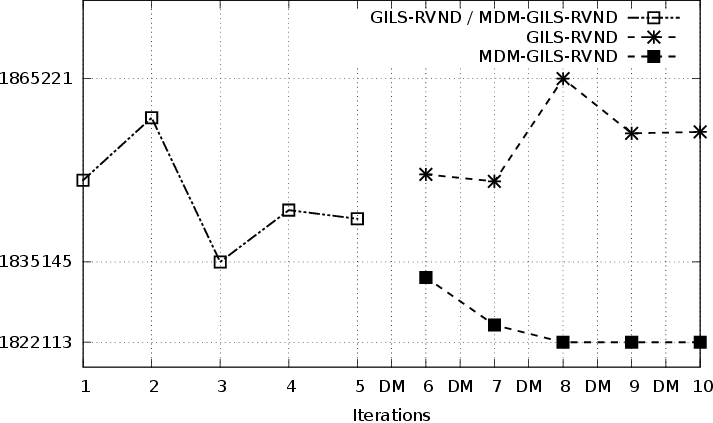}
		}
		\caption{Cost values versus Iteration - TRP-S500-R17 instance - 1st execution}
		\label{fig:behavior-r17}
		\centering
		\subfigure[Constructive phase]{
			\label{fig:constructive-r17-2}
			\includegraphics[scale=0.57]{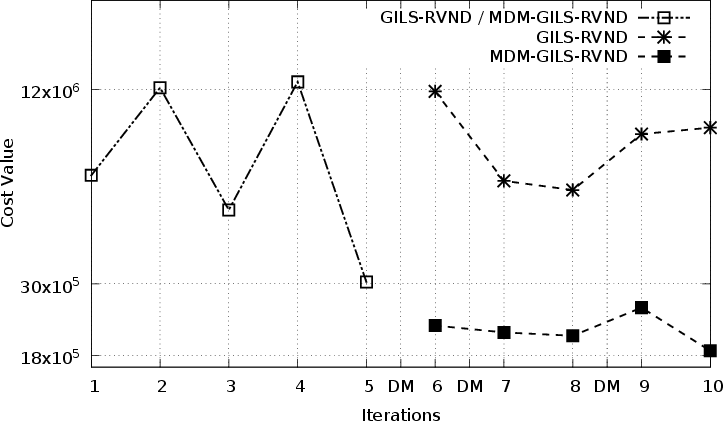}
		}
		\subfigure[Local search phase]{
			\label{fig:local-search-r17-2}
			\includegraphics[scale=0.57]{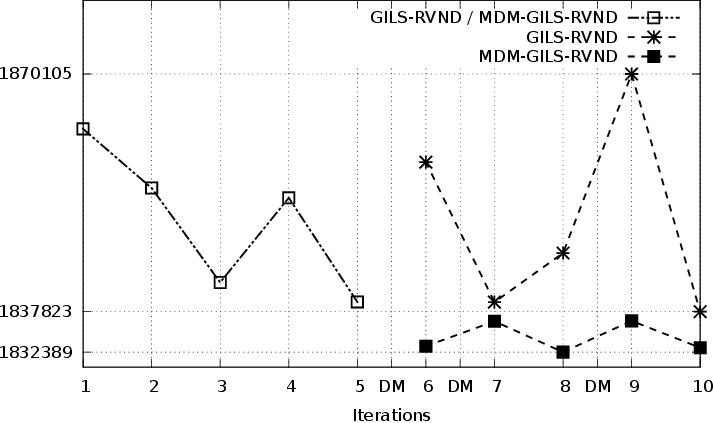}
		}
		\caption{Cost values versus Iteration - TRP-S500-R17 instance - 2nd execution}
		\label{fig:behavior-r17-2}
	\end{figure}

	To picture the behavior of a TRP-S500-R17 instance execution, Fig.~\ref{fig:constructive-r17} demonstrates that the obtained cost values from the data mining heuristic dropped dramatically due to the usage of a mined pattern, whereas the cost values from GILS-RVND roughly remain at the same baseline. Being specific at this point, each pattern used for constructing an initial solution has, on average, 480 frequent arcs. For the local search phase, displayed in Fig.~\ref{fig:local-search-r17}, from the first DM process onwards (between fifth and sixth iteration), the MDM-GILS-RVND always presented better cost values compared to their respective values obtained by GILS-RVND, including a progressive enhancement of the best solution found. In the second execution, significant reductions of the cost values from pattern-based solution were observed, illustrated in Fig.~\ref{fig:behavior-r17-2}. In this execution, as also demonstrated in Fig.~\ref{fig:constructive-r17}, MDM-GILS-RVND presented better cost values than the original heuristic in the two phases. As in the first execution, the average size of each pattern is 480 frequent arcs.

	\subsubsection{Analyses of Time Convergence} \label{subsubsec:time-convergence-analyses}
	
	In order to evaluate algorithms with random components, Time-to-target (TTT) plots are used to analyse their behaviors \citep{Aiex2007}. Particularly, a TTT plot displays the probability (ordinate axis) that algorithm will find a solution at least as good as a given target within a running time (abscissa axis).

	Four targets were chosen regarding three instances for the experiments, where the heuristics for both MLP versions were tested. The target 2672445 for the kroA200 instance and the target 6557983 for the pr299 instance were chosen based on the average solution obtained by GILS-RVND, as reported in Table~\ref{tab:mdm-results-circuit-new}. Targets 1834443 and 1830454 from the TRP-S500-R17 were chosen based on the average solution reached by, respectively, GILS-RVND and MDM-GILS-RVND, as reported in Table~\ref{tab:mdm-quality-s500}. For each target, 100 executions using distinct seeds of pseudo-random numbers were used in the experiments.
	
\begin{figure}[htbp]
		\centering
		\subfigure[2672445 as target for kroA200 instance]{
			\label{fig:hard-target-kroA200}
			\includegraphics[scale=0.58]{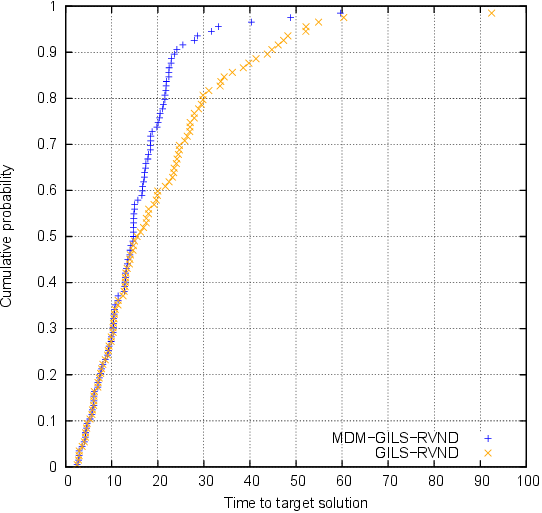}
		}
		\subfigure[6557983 as target for pr299 instance]{
			\label{fig:hard-target-pr299}
			\includegraphics[scale=0.58]{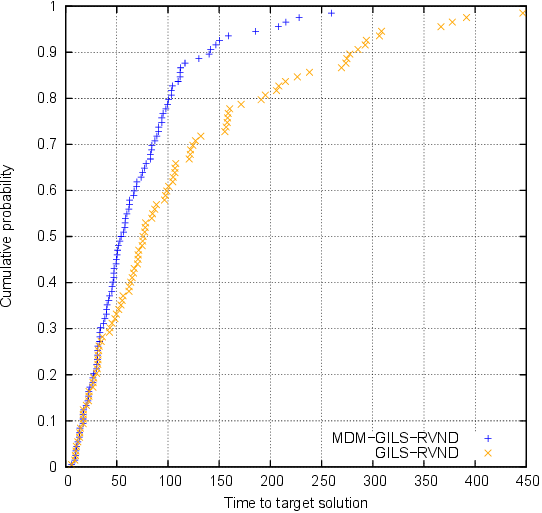}
		}
		\caption{TTT plots for  kroA200 and pr299 instances}
		\label{fig:tttplots-pr299}
		\subfigure[1834443 as target]{
			\label{fig:easy-target-S500-R17}
			\includegraphics[scale=0.58]{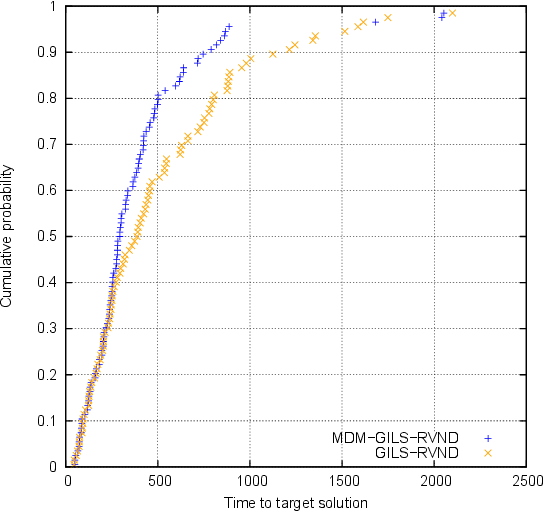}
		}
		\subfigure[1830454 as target]{
			\label{fig:hard-target-S500-R17}
			\includegraphics[scale=0.58]{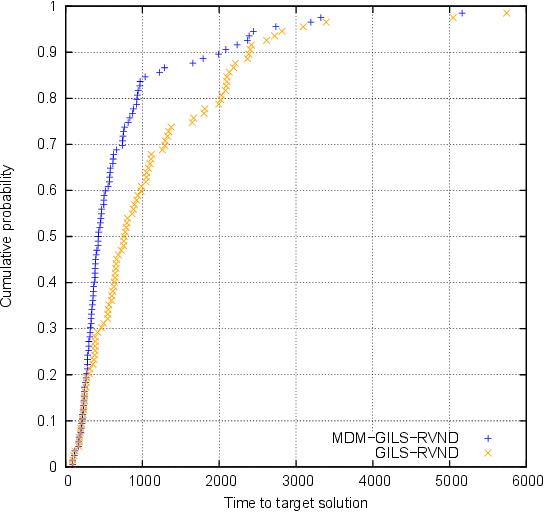}
		}
		\caption{TTT plots for TRP-S500-R17 instance}
		\label{fig:tttplots-S500-R17}
	\end{figure}

	Shown in Fig.~\ref{fig:hard-target-kroA200}, the plot using the target for kroA200 instance demonstrated distinct performances between both heuristics. For example, the GILS-RVND indicated a probability of around 65\% to reach the given target in 25 seconds, while MDM-GILS-RVND presented a probability of 92\% to reach the target within the same time. Looking at the TTT plot for the target from the pr299 instance, shown in Fig.~\ref{fig:hard-target-pr299}, it is clear to observe that MDM-GILS-RVND performed better than GILS-RVND for the given target. For example, the GILS-RVND strategy presented a probability around 80\% to attain the given target in 200 seconds, whereas the probability of MDM-GILS-RVND to reach the same target within the same time is about 95\%.
	
	Considering the TRP-S500-R17 instance, the figure involving the TTT experiments for the target 1834443 (easy target) is illustrated in Fig.~\ref{fig:easy-target-S500-R17}. In this scenario, the heuristics presented distinguishable behaviors. For instance, to attain the target in 500 seconds, the DM heuristic showed a probability of 81\%, whereas the original heuristic reached a probability of about 63\%. For the target 1830454 (hard target), which is shown in Fig.~\ref{fig:hard-target-S500-R17}, the heuristic with data mining presented a better convergence behavior to the target compared with the original strategy. As an example, the probability of MDM-GILS-RVND to achieve this target in 1000 seconds is about 85\%, while the probability of GILS-RVND is around 61\% to reach the given target within the same time.
	
	\subsubsection{Complementary Experiments} \label{subsubsec:complementary-experiments}
	
	In this subsection, a fair comparison between MDM-GILS-RVND and GILS-RVND is presented, using computational time as the stopping criterion. The longest computational time achieved between GILS-RVND and MDM-GILS-RVND, for the same seed of pseudo-random numbers, is given as input to the fastest heuristic for a new execution. This stopping criterion is verified after constructing an initial solution and after any complete neighborhood evaluation on the RVND method. It is noteworthy saying that such verifications have a negligible impact on the final computational time.
	
	For these experiments, we only report the results considering sets with challenging and relevant instances for the literature, which are the 56-instances set for the MLP of Hamiltonian circuits and the sets of 500 and 1000 customers, and the instance att532 of the 10-instance set selected by \cite{Salehipour2011} for the MLP of Hamiltonian paths. Results considering all instance sets and their execution logs are available on Mendeley~\cite[see][]{Santana2018dataset}. At the end of this subsection, all new BKSs found in our experiments for the MLP of Hamiltonian paths are reported in Table~\ref{tab:all-bks}.
	
	Results on the 56-instances set presented an overall improvement compared to the results reported in Table~\ref{tab:mdm-results-circuit-new}. In these new experiments, the DM strategy outperformed, even more, the results of GILS-RVND, where, for the best solution and average solution terms, MDM-GILS-RVND achieved, in both aspects, 51 better results. Considering statistical significance tests, the depth of enhancement is verified by the number of results statistically significant. Indeed, 16 results for the Student's t-test, and four results for the Wilcoxon test were statistically significant when MDM-GILS-RVND performed better than GILS-RVND.

	Considering the 500-customers set, the DM heuristic showed a better behavior compared to GILS-RVND, where 19 best average solutions and 19 best solutions were improved. Comparing their gaps, represented by the averages from the BKS gap columns, 0.51\%, and 0.33\% were obtained, respectively, by GILS-RVND and MDM-GILS-RVND. In statistical significance terms, 11 results using Student's t-test were statistically significant when MDM-GILS-RVND had better performances over GILS-RVND. New BKS were obtained in 11 instance results by MDM-GILS-RVND.

	For the experiments involving the 1000-customers set, considering the best solution and average solution aspects simultaneously, the DM strategy outperformed GILS-RVND in all 20 instances of this set. Averages from the BKS gap columns were 0.57 and 0.15, respectively, for GILS-RVND and MDM-GILS-RVND. All 20 instance results using Student's t-test were statistically significant when MDM-GILS-RVND outperformed GILS-RVND. Finally, all 20 best solutions obtained by MDM-GILS-RVND improved the existing BKSs of this instance set.
	
	\begin{table}
		\centering
		\caption{32 new best known solutions achieved for several instances of the Minimum Latency Problem}
		\label{tab:all-bks}
		\resizebox{\textwidth}{!}
		{
			\begin{tabular}{lr|lr||lr|lr}
				Instance     &  New BKS     & Instance     &  New BKS &     Instance     &  New BKS &     Instance &     New BKS  \\\hline
				att532       & 5572131& TRP-S500-R15 & 1784919&  TRP-S1000-R5 &  5085412  &  TRP-S1000-R13&  5083179    \\
				TRP-S500-R1  & 1841210& TRP-S500-R17 & 1819909&  TRP-S1000-R6 &  5087134  &  TRP-S1000-R14   &    5076253    \\
				TRP-S500-R3  & 1826738&  TRP-S500-R19 & 1776855&   TRP-S1000-R7 &  4980214  &  TRP-S1000-R15   &     5121456   \\
				TRP-S500-R4  & 1802921&   TRP-S500-R20 & 1820168&  TRP-S1000-R8 &  5096892  &  TRP-S1000-R16   &   5041659     \\
				TRP-S500-R7  & 1846251&  TRP-S1000-R1 & 5097334&  TRP-S1000-R9 &  5022622   &  TRP-S1000-R17   &     5045873   \\
				TRP-S500-R9  & 1729796&   TRP-S1000-R2 &  5082922   &  TRP-S1000-R10&  5053780   &  TRP-S1000-R18   &  4984408   \\
				TRP-S500-R11 & 1797771&   TRP-S1000-R3 &  5080369    &  TRP-S1000-R11&  5022026  &  TRP-S1000-R19   &   5058047 \\     
				TRP-S500-R14 & 1796129&   TRP-S1000-R4 &  5092276    &  TRP-S1000-R12&  5004216   &  TRP-S1000-R20   &   4964371     \\
			\end{tabular}                
		}
	\end{table}

	\section{Conclusion and Future Works} \label{sec:conclusion}
	
	In this paper, a hybrid heuristic using data mining techniques was conceived based on of a state-of-the-art heuristic (GILS-RVND) for two Minimum Latency Problem (MLP) variants. 
	
	This hybrid heuristic, named MDM-GILS-RVND, consists in an adapted version of the classic MDM-GRASP (Multi DM-GRASP) that performs the data mining process whenever the elite set gets updated. In this strategy, patterns found in high-quality solutions are used to generate initial solutions.
	
	In order to provide a fair and strict evaluation of the computational experiments, all GILS-RVND experiments reported in~\cite{Silva2012} were entirely re-executed. Additionally, 38 instances selected from TSPLib were introduced to the experiments of the MLP version of Hamiltonian circuits. In total, 229 instances were tested in this paper, being 79 and 150 instances, respectively, for the MLP version of Hamiltonian circuits and Hamiltonian paths.
	
	Moreover, analyses of time convergence (TTT plots), examples of the impact of mined patterns into initial solutions, and statistical significance tests were done to support the evaluation of the heuristics. Computational experiments using running time as stopping criterion were also carried out.
	
	Reported results demonstrated that as the problem gets harder (i.e., the number of customers increases), the more efficient the data mining heuristic performed. Considering the results obtained from the reproduced experiments, MDM-GILS-RVND proved to perform better than GILS-RVND in terms of solution quality and computational time simultaneously. For example, in harder instances (with 500 customers or more) for both MLP variants, we observe that the improvements of solution quality made by the data mining heuristic were significant since their numbers of wins over GILS-RVND were predominant. Regarding the experiments based on computational time as a stopping criterion, when both heuristics are executed within the same amount of time, the DM heuristic reached far better results.
	
	It is worth to mention that 32 new BKSs were achieved by MDM-GILS-RNVD for the MLP based on Hamiltonian paths. Moreover, all tested instances used in this work and numerical results are available in an online repository on Mendeley \citep[see][]{Santana2018dataset}.
	
	A future improvement to be further investigated consists in employing a different application of mined patterns. In \cite{Maia2016}, a mined pattern is used to reduce the problem's size rather than generating initial solutions. In the context of MLP, each segment of solution in a given pattern could be contracted to a single vertex. Then, the heuristic solves the reduced instance, and the obtained solution is expanded in order to reach a solution to the original instance.
	
	\section*{Acknowledgments} This research was partially supported by the Brazilian agencies CNPq and CAPES.  
	
	\nocite{*}
	
	\bibliographystyle{itor}
	\bibliography{main}
	
\end{document}